\pdfoutput=1

\documentclass[11pt]{article}

\usepackage[final]{acl}

\usepackage{times}
\usepackage{latexsym}

\usepackage[T1]{fontenc}

\usepackage[utf8]{inputenc}

\usepackage{microtype}

\usepackage{inconsolata}

\usepackage{graphicx}

\usepackage{amsmath,amssymb,amsthm}
\newtheorem{theorem}{Theorem}[section]
\newtheorem{lemma}[theorem]{Lemma}
\newtheorem{corollary}[theorem]{Corollary}
\newtheorem{definition}[theorem]{Definition}

\newtheorem{remark}[theorem]{Remark}

\providecommand{\R}{\mathbb{R}} %

\providecommand{\ff}{\mathbf{f}}
\let\ggg\gg
\renewcommand{\gg}{\mathbf{g}}

\providecommand{\zz}{\mathbf{z}}

\providecommand{\mLambda}{\boldsymbol{\Lambda}}

\providecommand{\cA}{\mathcal{A}}

\providecommand{\cM}{\mathcal{M}}

\newenvironment{talign*}
{\csname align*\endcsname}
{\endalign}

\usepackage[utf8]{inputenc}         %
\usepackage[T1]{fontenc}            %
\usepackage{url}                    %
\usepackage{booktabs}               %
\usepackage{amsfonts}               %
\usepackage{nicefrac}               %
\usepackage{microtype}              %
\usepackage{xcolor}                 %
\usepackage{algorithm}
\usepackage{algpseudocode}
\usepackage{graphicx}
\usepackage{subcaption}
\usepackage[flushleft]{threeparttable}
\usepackage{float}
\usepackage{multirow}
\usepackage{xspace}
\usepackage{natbib}
\usepackage{enumitem}
\usepackage[font=small]{caption}
\usepackage{autobreak}
\usepackage{sidecap}
\usepackage{wrapfig}
\usepackage{bbding}
\usepackage[toc, page, header]{appendix}
\usepackage{tikz}
\usepackage{xcolor}
\usepackage{pifont}
\usepackage{mdframed}
\usepackage{hyperref}
\usepackage{makecell}

\hypersetup{
    colorlinks=true,
    linkcolor=blue,
    citecolor=blue,
    urlcolor=blue
}

\definecolor{coral}{RGB}{255,127,80}
\definecolor{darkgreen}{RGB}{0,100,0}
\definecolor{darkyellow}{RGB}{204,153,0}
\definecolor{salmon}{RGB}{250,128,114}

\algrenewcomment[1]{\textcolor{blue}{\hfill// #1}}

\usepackage{xspace}

\usepackage[textwidth=3.5cm]{todonotes}

\usepackage{float}

\title{Keys to Robust Edits: From Theoretical Insights to Practical Advances}

\author{%
Jianhao Yan$^{1,2}$\thanks{~~Two authors contributed equally to this work.} \hspace{1.5em}
Futing Wang$^{1,2}$\footnotemark[1] \hspace{1.5em}
Yun Luo$^{1,2}$ \hspace{1.5em}
Yafu Li$^{4}$ \hspace{1.5em}
Yue Zhang$^{2,3}$\thanks{~~Corresponding author.} \\
\centerline{\normalfont{$^1$Zhejiang University} \quad \normalfont{$^2$School of Engineering, Westlake University}} \\
\centerline{\normalfont{$^3$ Institute of Advanced Technology, Westlake Institute for Advanced Study}} \\
\centerline{\normalfont{$^4$ Shanghai AI Lab}} \\
\centerline{\texttt{elliottyan37@gmail.com}}
}

\begin{document}
\maketitle

\setlength{\abovedisplayskip}{5pt}
\setlength{\belowdisplayskip}{5pt}
\setlength{\abovedisplayshortskip}{5pt}
\setlength{\belowdisplayshortskip}{5pt}

\begin{abstract}
Large language models (LLMs) struggle with maintaining accurate knowledge due to conflicting/outdated parametric memories. While locate-and-edit methods address this, their reliance on models' internal representations leads to robustness failures in long-context reasoning and paraphrased queries. We identify a fundamental limitation of locate-and-edit methods: existing semantic keys (for memory localization) cannot simultaneously satisfy robustness (context-invariant activation) and specificity (precise knowledge discrimination). Through theoretical error-bound analysis, we establish formal criteria for effective editing.
Our solution introduces \textit{Robust Edit Pathway (REP)}, a plug-and-play module that: (1) disentangles editing keys from native model representations; (2) dynamically adjusts keys via contrastive learning to achieve robustness-specificity balance. 
Extensive experiments across various editing methods (ROME/MEMIT/R-ROME/EMMET), existing LLMs (LLaMA2, QWen, Mistral), and datasets (CounterFact, ZsRE) show that REP improves success rate over robustness tests by up-to 66.4\% while maintaining the success rate unaffected. 
\footnote{Our code can be found at \url{https://github.com/ElliottYan/RobustKeyEdit}.}

\end{abstract}

\section{Introduction}

Large language models (LLMs, \citealt{achiam2023gpt,llama,llama2}) have revolutionized knowledge storage through their parametric memories, yet their reliance on static training data renders them prone to inaccuracies from conflicting or outdated information. 
While knowledge editing methods like ROME and MEMIT~\cite{meng2022locating,meng2022mass} attempt to address this by modifying specific model parameters, existing approaches are found to suffer from editing failures with robustness tests~\cite{ma2024possible,yang2024butterfly}. 
For example, editing "Slovenia belongs to Europe → Antarctica" frequently collapses when the subject is rephrased ("Republic of Slovenia"), embedded in long contexts, or attacked by shuffling subjects.
The unreliability greatly limits the impact and application of model editing methods.

We uncover a fundamental flaw in their core mechanism: \emph{the intrinsic instability of the model's internal representations when used as semantic keys for editing. }
Existing approaches assume these internal representations can reliably localize knowledge.
Through formal analysis of key-value associative memory in MLP layers (Definitions \ref{def:kv}  - \ref{def:rome_solution}) and empirical analyses, we prove that existing internal representations frequently violate the foundational conditions for reliable editing:
(1) \textbf{Key Sensitivity}: Representations of the same fact diverge drastically under perturbations. Whitened similarity scores drop to near-random levels for shuffled subject tokens (e.g., "\_ia Sloven" vs. "Sloven \_ia") and for rephrased variants, breaching the robustness bound derived in Lemma \ref{lemma:generalize}; (2) \textbf{Key Collisions}: Semantically distinct entities exhibit unintended overlaps in the whitened space for unrelated pairs like "Michael Jordan" and "Kobe Bryant", Figure \ref{fig:unrelated}), contradicting the specificity requirement in Lemma \ref{lemma:spec}.  

\begin{figure*}[t]
    \centering
    \includegraphics[width=0.86\textwidth]{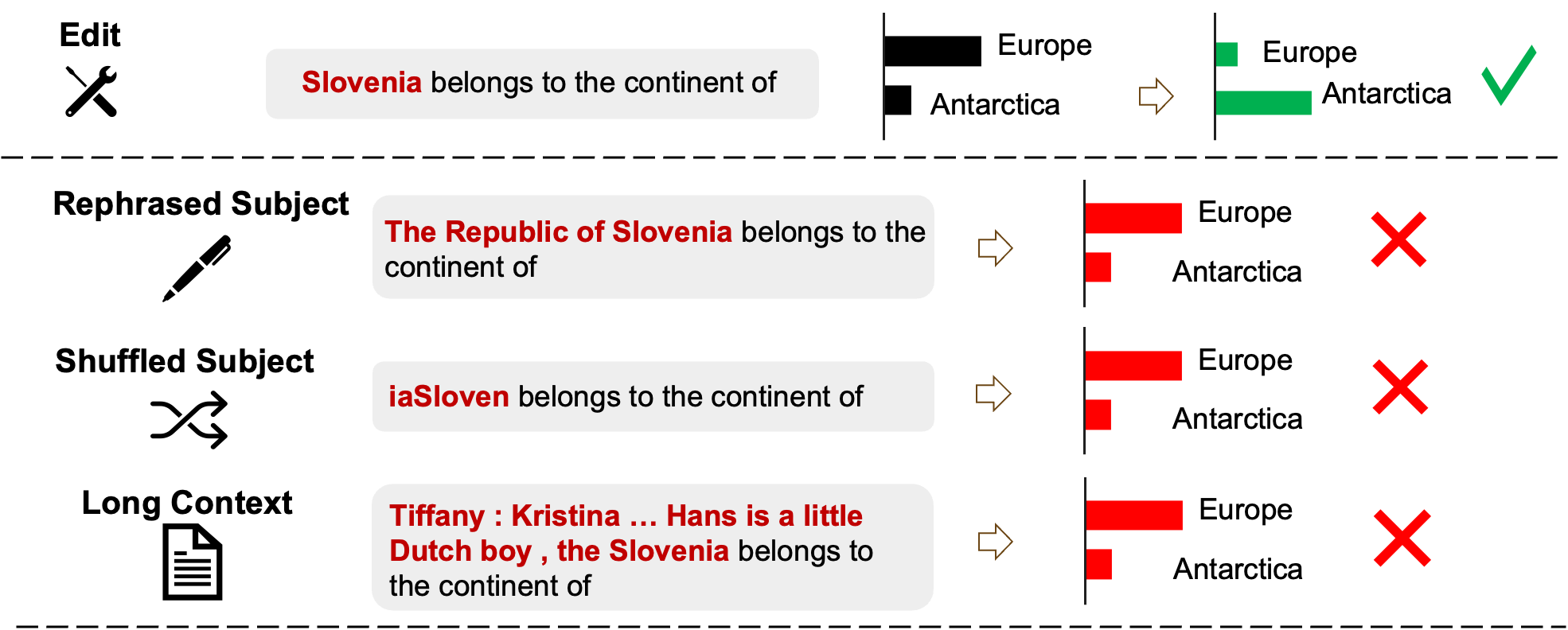}
    \caption{An example of the edited knowledge `Slovenia belongs to the continent of' through knowledge editing and its failures on the different scenarios.}
    \vspace{-20pt}
    \label{example}
  \end{figure*}

To resolve this, we propose \textbf{Robust Edit Pathway (REP}), a novel plug-and-play module that disentangles editing keys from native model representations. 
Inspired by our theoretical results, 
where effective knowledge insertion requires 
both centering around semantically equivalent surface forms of subjects while not affecting unrelated ones -- REP introduces: (1) Disentangled Key Projection: A contrastively trained adapter aligns keys for target facts across perturbations, ensuring context-invariant activation through whitened similarity optimization (Eq. \ref{eq:final_training}); (2) Dynamic Gate Mechanism: Token-level gating selectively activates edits, dynamically balancing robustness and specificity.
Extensive evaluations across 4 editing methods (ROME/MEMIT/R-ROME/EMMET)\citep{meng2022locating, meng2022mass, gupta2024rebuilding, yoon2024bigger}, 3 LLMs~(LLaMA2-7B, Mistral-7B, and Qwen-2-7B)\citep{llama2, jiang2023mistral, yang2024qwen2}, and two datasets \citep{meng2022locating, de-cao-etal-2021-editing} demonstrate REP’s superiority: (1) up-to 66.4\% absolute gains on robustness tests, recovering at most 94\% of the editing performance versus unperturbed inputs; (2) specificity preservation ($\Delta$Locality < 1.6) and minimal fluency degradation ($\Delta$Fluency <2.2); (3) effectiveness on both in-domain and out-of-domain robustness queries.

Our contributions are as follows: 
\begin{itemize}
    \item Through theoretical error-bound analysis, we establish formal criteria for effective model editing and reveal fundamental limitations in using internal representations as editing keys.
    \item Extensive experiments demonstrate existing semantic keys cannot simultaneously achieve robustness (context-invariant activation) and specificity (precise knowledge discrimination).
    \item We propose Robust Edit Pathway (REP), a plug-and-play module that disentangles editing keys from native model representations and dynamically adjusts them via contrastive learning.
    \item Experiments across various editing methods (ROME/MEMIT/R-ROME/EMMET), LLMs, and datasets show REP improves success rate over robustness tests by up-to 66.4\% while maintaining editing performance.
\end{itemize}

\section{Related Work}
\paragraph*{Knowledge Editing.} As large language models have grown in complexity and size, post-modification has become increasingly challenging due to their opaque mechanisms and vast parameter spaces~\citep{mitchell2022memory,zhong2023mquake}. This has led to heightened interest in knowledge editing, a technique for precise model modification. 
Knowledge editing are applied to various scenarios, such as editing for safety~\citep{wang2024detoxifying}, debias~\citep{yan2024potential} and concepts~\citep{wang2024editing}. 


Our work is in line with the \emph{locate-and-edit} methods, which draw much attention as they potentially unveil how the knowledge are stored in an LLM. 
These approaches first identify relevant parameters before updating them to modify specific knowledge, including KnowledgeNeuron's attribution-based neuron updating \citep{dai2021knowledge}, ROME's causal mediation analysis for MLP editing \citep{meng2022locating}, MEMIT's multi-layer residual distribution \citep{meng2022mass}, PMET's refined allocation strategy \citep{li2024pmet}, and WilKE's dynamic layer selection \citep{hu2024wilke} to reduce potential negative effects.
These methods all utilize inner representations as keys for key-value modeling. 
In contrast, we show that inner representations cannot meet the requirements of robust and specific edits, and we propose a robust edit pathway to mitigate this. 


\paragraph*{Challenges of Knowledge Editing.} 
Despite the promise, various challenges persist in practical applications of model editing methods. Previous studies show that edits often degrade general language abilities \citep{gu2024model,ma2024perturbation}, damage the hidden space \citep{wang2024missing}, struggle to propagate to related facts \citep{hua2024propagation}, and are easily forgotten during sequential updates \citep{gupta2024model}. Moreover, multi-hop reasoning can elicit old knowledge \citep{zhang2024enhancing}, and models may collapse after few edits \citep{yang2024butterfly,brown2023edit}.

Further complications include cross-lingual inconsistencies \citep{wang2024crosslingualknowledgeeditinglarge}, knowledge conflicts \citep{li2023unveiling}, and inadequate evaluation in realistic settings such as long-form generation \citep{rosati2024long} and neighborhood knowledge \citep{ma2024neighboring}. 
These issues underscore the need for more sophisticated and comprehensive editing techniques. 
However, previous research largely remains focused on the outcomes of knowledge editing in various scenarios, lacking a deeper understanding of the underlying mechanisms of these methods and the true reasons behind their frequent failures. 
Our work presents both theoretical and empirical understanding regarding the reason for the robustness failures of locate-and-edit methods and proposes REP to enhance them. 


\begin{figure*}
    \begin{minipage}[t]{0.6\linewidth}
      \vspace{0pt}  
      \centering
      \includegraphics[width=\linewidth]{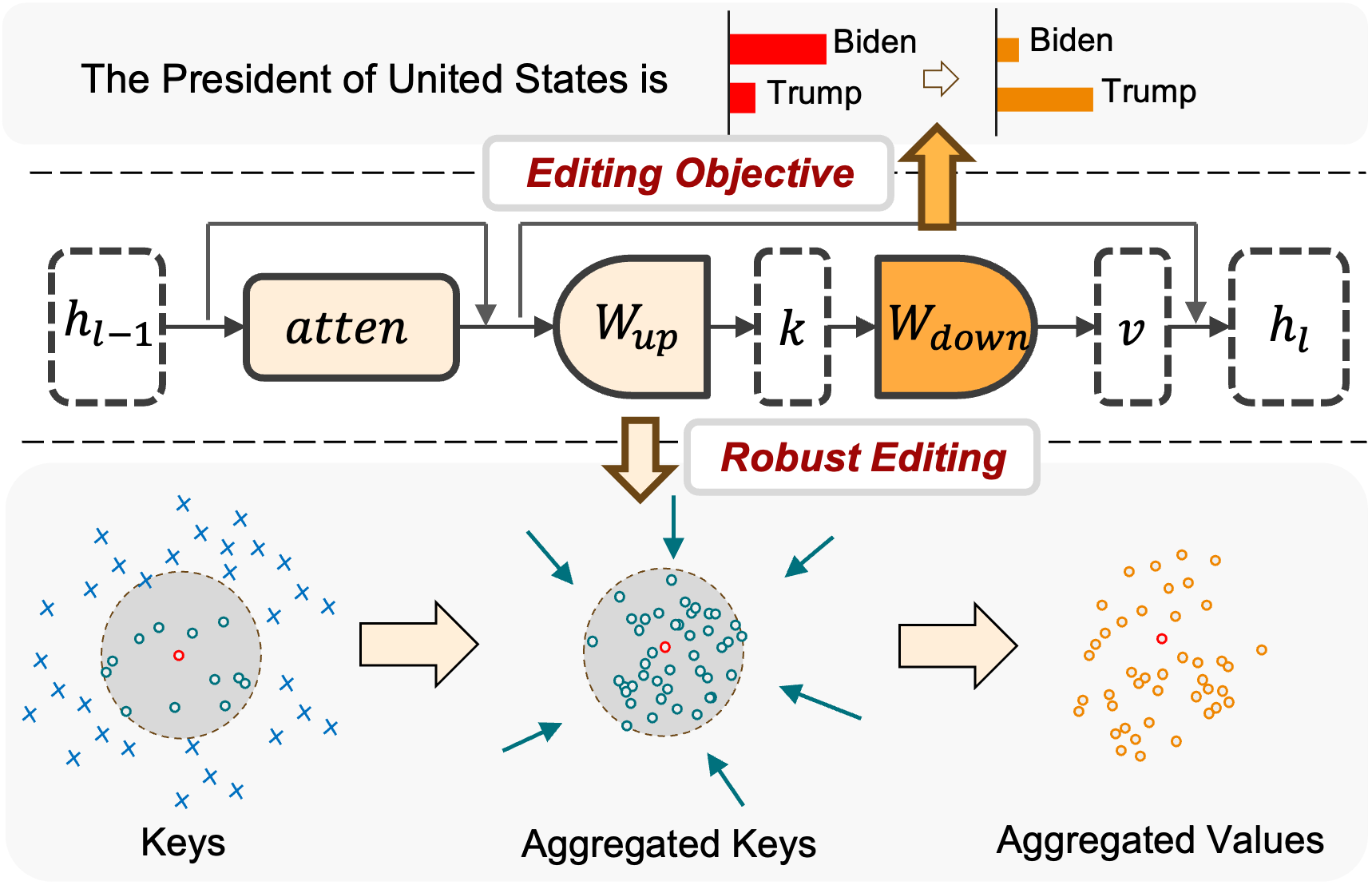}
      
    \end{minipage}
    \hfill
    \begin{minipage}[t]{0.33\linewidth}
      \vspace{0pt}  
      \centering
      \includegraphics[width=\linewidth]{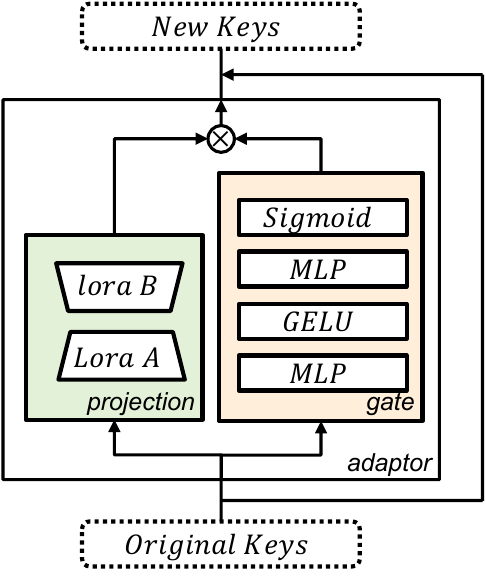}
    \end{minipage}
    \vspace{{-5pt}}
    \caption{Overview of REP. \textbf{Left}: Key concept visualization; \textbf{Right}: Architectural design of the adapter.}
    \label{fig:overall}
    \vspace{{-15pt}}
\end{figure*}

\section{Knowledge Editing}
In this section, we first formulate 
knowledge editing and review the \emph{locate-and-edit} methods. 
\paragraph{Task Definition} Knowledge editing focuses on updating factual associations in language models. 
Following \cite{meng2022locating} and \cite{meng2022mass}, we define a knowledge $\ff$ as a triple $(h, r, t)$, where $h$ is the head entity, $r$ is the relation, and $t$ is the target entity (e.g., (USA, {has president}, Biden)).
Given a knowledge triple: $(h, r, t)$, the goal is to modify the model's knowledge by replacing the target entity $t$ with a new target $t_*=\text{Trump}$ (e.g., changing `Biden' to `Trump').

Autoregressive large language models (LLMs) can {complete a natural-language sentence} by leveraging implicit knowledge encoded within their parameters. 
Thus, a knowledge triple $(h, r, t)$ is considered stored in the LLM when the model can predict the target $t$ given a prompt that corresponds to $(h, r, \cdot)$.
For instance, given a prompt `The president of USA is', a model with the above knowledge would predict `Biden'.

\begin{definition}[\textbf{Knowledge Editing for LLMs}]
    Given a knowledge triple $(h, r, t)$ already stored in the language model $\cM$ and a new knowledge $(h, r, t_*)$, there exists a set of prompts $P=\{p\}$ corresponds to $(h, r, \cdot)$.
    The knowledge editing algorithm $\cA$ aims to modify the model's prediction on $P$ from $t$ to $t_*$. This task can be formally expressed as follows:
    \begin{gather*}
        \cM' = \cA(\cM) \,, \\\text{s.t.} ~ \cM(p) = t,  \cM'(p) = t_*, \forall p \in P,
    \end{gather*}
\end{definition}


\paragraph{Architectural Foundations for locate-and-edit}
The efficacy of \emph{locate-and-edit} methods relies on identifying modular components in LLMs that encode factual knowledge. Transformer-based ~\cite{vaswani2017attention} LLMs organize computation into layers containing two core submodules: self-attention (for contextual reasoning) and Multi-Layer Perceptrons (MLPs, for nonlinear feature transformations). A key insight from ROME \citep{meng2022locating} establishes that factual associations localize to specific MLP layers—enabling precise edits.

Each MLP layer comprises two feed-forward operations: (1) an \textit{up-projection} that expands hidden dimensions for fine-grained feature interactions, and (2) a \textit{down-projection} that contracts dimensions to synthesize higher-level representations. ROME treats these MLPs as linear associative memories (Definition \ref{def:kv}), leveraging causal mediation analysis to pinpoint layers where edits (e.g., substituting “Biden” → “Trump” in presidential facts) propagate correctly. By surgically modifying these layers, ROME updates targeted knowledge while preserving unrelated model capabilities, minimizing unintended side effects.



\begin{definition}[\textbf{MLP Layers as Associative Memories}]\label{def:kv}
    The down-projection weight matrix $W$ in the MLP layer can be interpreted as a linear associative memory system. Specifically:
    \begin{itemize}[leftmargin=12pt,nosep]
        \item Keys $K = [k_1 | k_2 | \cdots | k_n] \in \mathbb{R}^{D_1 \times n}$ represent the intermediate representations of the prompt corresponding to $(h, r,\cdot)$ before down-projection.
        \item Values $V = [v_1 | v_2 | \cdots | v_n] \in \mathbb{R}^{D_2 \times n}$ represent the corresponding outputs after down-projection.
    \end{itemize}
    The weight matrix $W \in \mathbf{R}^{D_2\times D_1}$ approximately maps the keys to their associated values, satisfying $W K \approx V$.
    
\end{definition}
Definition \ref{def:kv} (illustrated by
Figure~\ref{fig:overall} left) enables the MLP layers to store and retrieve prompt-target associations.
Then, ROME accomplishes knowledge editing by inserting a new key-value pair into the MLP layer, modifying $W$ to ${\hat{W}}$.
\begin{definition}[\textbf{The Solution of ROME}] \label{def:rome_solution}
    
    In ROME, a new key-value pair $(k_*, v_*)$ can be inserted into the language model using the following closed-form solution:
    \begin{gather*}
        \text{minimize}_{\hat{W}}\ ||\hat{W}K - V||
        ~\text{s.t.} \ \hat{W}k_* = v_* \,, \\ \text{by setting} \ \hat{W} =  W + \mLambda(C^{-1}k_*)^T
    \end{gather*}
    where:
    \begin{itemize}[leftmargin=12pt,nosep]
        \item $C=KK ^T$ is a constant matrix pre-cached by estimating the uncentered covariance of $k$ from a sample of Wikipedia text,
        \item $\mLambda=\frac{v_*-W k_*}{(C ^{-1}k_*)^T k_*}$ is a vector proportional to the residual error of the new key-value pair on the original memory matrix.
    \end{itemize}
\end{definition}

{Intuitively, $||\hat{W}K - V||$ controls the shift from previously stored keys and values, and $\hat{W}k_* = v_*$ makes sure that the new knowledge is added into $\hat{W}$.} To implement this solution, it is necessary to extract the key $k_*$ and calculate the value $v_*$.

\begin{remark}[\textbf{Extract $k_*$}] 
    In $\cM$, $k_*$ is obtained by averaging the activations collected at the last token of the head entity $h$, processing a small set of texts that end with the head entity $h$. This can be formally written as:
    \begin{align*}  
        k_* = \frac{1}{M} \sum^M_{{j=1}} k(x_j + h)\,,
    \end{align*}
    where $k(\cdot)$ is the input of the second MLP layer of the $l_*$-th FFN layer in the transformer, $M$ is the number of the selected texts and $x_j$ represents a random prefix.
\end{remark} 

Once $k_*$ is extracted, the next step is to determine the appropriate value $v_*$ for the new key-value pair.

\begin{remark}[\textbf{Calculate $v_*$}]
    Let $\mathbb{P}_{\cM'}(t_*|p)$ denote the probability of $t_*$ after $\cM$ processes query prompt $p$. We seek a vector $\zz$ to substitute as the output of the MLP in layer $l^*$ at token $i$ (denoted $m_{i}^{(l*)}:\zz$) such that the network predicts the target tail entity $t_*$ while maintaining the model's understanding of the subject's essence. The optimization objective is as follows:
    \begin{align*}
        v_* = \arg\min_\zz \frac{1}{N}\sum_{j=1}^{N}\underbrace{-\log\mathbb{P}_{\cM(m_{i}^{(l*)}:\zz)}[h'|x_{j}+p]}_{\text{(a) Maximizing } h' \text{ probability}} \\
        +\underbrace{D_{\text{KL}}\left(\mathbb{P}_{\cM(m_{i}^{(l*)}:\zz)}[\cdot|p']||\mathbb{P}_{\cM}[\cdot|p']\right)}_{\text{(b) Controlling essence drift}}.
    \end{align*}
    where $p'$ is `{subject} is a'.
\end{remark}

In conclusion, the ROME method effectively enables the insertion of new knowledge triples $(h, r, t_*)$ into large language models through operating key-value pairs.

\section{Theoretical Results of Key-Value Associative Memory}
The idea of keys and values in associative memory (as shown in Definition~\ref{def:kv}) is analogous to the key-value databases in modern computer systems.
What makes the difference here is that down-projection FFNs implement a fuzzy retrieval mechanism, whereas modern key-value databases generally require the keys to be unique.
\begin{lemma}[\textbf{Fuzzy Key-Value Mapping}]
    Given $K\in\R^{D_1 \times n}$ and $V\in\R^{D_2 \times n}$ as defined in Definition \ref{def:kv} that are already stored in the feed-forward layer $W\in\R^{D_2\times D_1}$, assume $n \ggg D_1$ and $K$ has the rank of $D_1$. 
    When a new query $k_*$ comes, its corresponding value can be represented as the weighted sum of existing values, $v_* = \sum^N_{{i=0}} \alpha_i v_i$ and $\alpha = K^T(KK^T)^{-1}k_*$ can be solved by the Moore-Penrose pseudoinverse.
    \label{lemma:fuzzy}
\end{lemma}
Lemma \ref{lemma:fuzzy} demonstrates that the retrieved memory of a new test query can be considered as the linear combination of previously stored memory, which leads to the following direct corollary. 

\begin{corollary}[\textbf{Edited Key-Value as a Patch against Original Knowledge}]
    In locate-and-edit algorithms, new knowledge is injected into the memory as a key-value pair $(k_*, v_*)$. Consider a set of existing key-value pairs $(k_i, v_i)$ where $k_i \in \mathcal{K}_s, v_i\in V_s$ that represent the same knowledge as $(k_*, v_*)$ (e.g., paraphrases). 
    Suppose the injection is lossless\footnote{In a real-world scenario, the edit cannot be lossless. Here, for a clear intuition, the above lemma is presented in an ideal way as the editing process will change the value of previously stored key-value pairs. We show that even considering the lossless scenario, the current LLMs cannot satisfy robustness and specificity requirements.} 
    and that $\mathcal{K}_s$ has full row rank, querying with any $k_i \in \mathcal{K}_s$ would retrieve a value $v=\sum_{i} \alpha_i v_i + \alpha_* v_*$.
\end{corollary}
\begin{remark}
    This corollary reveals that knowledge editing operates as an additive mechanism rather than a replacement one.
    Instead, it leaves the previously stored knowledge intact and counters them with a newly added value $v_*$.
\end{remark}

\begin{lemma}[\textbf{Bound on optimized $\Delta v=v_* - v_o$}]
    Assume the edited layer is only connected to the final prediction layer via an attention layer, where the attention layer has parameters $W_Q, W_K, W_V$, and $w_{t}$ and $w_{t_*}$ are the output embeddings for the original and edited target, we have the following inequality, 
\begin{align*}
    & (w_{t_*} - w_t)^T W_V (v_* - v_o) > \epsilon_1 + \epsilon_2 \\
    ~\Rightarrow & || (w_{t_*} - w_t)^T W_V || \cdot ||v_*-v_o|| >  \epsilon_1 + \epsilon_2,
\end{align*}
where $\epsilon$ is the logit gap after projection to the output embedding between $t$ and $t_*$. $\epsilon_1$ denotes the logit gap before edit and $\epsilon_2$ denotes the logit gap after edit. 
A value of $\epsilon\approx 2.30$ corresponds to a 90\% top-1 prediction probability. 
\label{lemma:value_bound}\end{lemma}
\begin{remark}
    Lemma \ref{lemma:value_bound} suggests that an edited value should be first similar to the vector pointing from $t$ to $t_*$ after a projection with $W_V$. Then, the edited values $\Delta v$ should be sufficiently large to ensure the success of the edit. 
\end{remark}
Our assumption here simplifies the connection between the edited layer and the prediction layer, as in real-world scenarios, the edit layer might pass through subsequent layers and undergo multiple attention operations before finally connecting to the prediction layer. 
However, the path we're considering~(i.e., from the edit layer to the prediction layer via an attention layer) is arguably the most direct route. 
We contend that this direct path is crucial and warrants particular attention, and this simplification allows us to focus on the most immediate and potentially significant impact of edits. 

\begin{lemma}[\textbf{Robustness Requirement for the Key-Values}]
    Robust editing requires consistency across semantically equivalent inputs: when editing knowledge with a new pair $(k_*, v_*)$, the edit should propagate to all semantically equivalent representations in the memory. For an edit to be considered a robust edit, querying with any $k_s \in \mathcal{K}_s$ should reliably retrieve the new knowledge $(h, r, t_*)$. This can be expressed as the following constraint:
    \begin{equation*}
         (w_{t_*}-w_{t})^T W_V (k_s^T C^{-1} k_*) \cdot v_*^T > \epsilon_1 + \epsilon_2, \forall k_s \in \mathcal{K}_s.
    \end{equation*} 
    \label{lemma:generalize}
\end{lemma}
When we look into the Lemma \ref{lemma:generalize}, $\beta_{s, *}=k_s^T C^{-1} k_*$ can be seen as a similarity measure on a projected space, namely whiten similarity. 
This lemma implies that (1) $v_*$ is decided by $\min_{k_{s'} \in \mathcal{K}_s} (k_{s'} C^{-1} k_*$), that is, $k_*$ should be near all $k_s \in \mathcal{K}_s$. If not, $v_*$ needs to be of large magnitude to counter the difference. Such large-magnitude updates can destabilize the model's learned representations and potentially degrade its overall performance; (2) $v_*$ needs not only to be aligned with the direction $(w_{t_*}-w_{t})$, but also has a sufficiently large magnitude to ensure editing success. 

\begin{lemma}[\textbf{Specificity Requirement for the Key-Values}]
If the newly added knowledge triplet $(h, r, t_*)$ would not be retrieved for any $k_o \notin \mathcal{K}_s$, it requires the following inequality to be satisfied:
    \begin{gather*}
        (w_{t_n}-w_{t_*})^T W_V (k_o^T C^{-1} k_*) \cdot v_*^T < \epsilon_3, \\
        \forall k_o \notin \mathcal{K}_s~and~\forall w \in W,
    \end{gather*}
    where $t_n$ is the original target retrieved by $k_o$ and $\epsilon_3$ denotes the logit difference between $t_n$ and $t_*$.
    \label{lemma:spec}
\end{lemma}
One simple solution for this lemma is $k^T C^{-1} k_* = 0$, which describes no superposition, as discussed in one of the concurrent work~\citep{hu2024knowledge}. However, as superposition generally exists among existing LLMs, we discuss more general cases here. 
\begin{lemma}[Whitened Similarity Bounds]
For a successful edit to achieve both robustness and specificity, the whitened similarities must satisfy:
\begin{enumerate}
    \item Lower bound for semantically equivalent keys:
    \begin{equation}
        \beta_{s,*} = k_s^T C^{-1}k^* \geq \beta_{\min}, \quad \forall k_s \in \mathcal{K}_s
    \end{equation}
    where $\beta_{\min} = \frac{\epsilon_1 + \epsilon_2}{||(w_{t^*} - w_t)^T W_V|| \cdot ||v^*||}$
    
    \item Upper bound for unrelated keys:
    \begin{equation}
        |\beta_{o,*}| = |k_o^T C^{-1}k^*| \leq \beta_{\max}, \quad \forall k_o \notin \mathcal{K}_s
    \end{equation}
    where $\beta_{\max} = \frac{\epsilon_3}{\max_{w \in \mathcal{W}} ||w - w_{t^*}|| \cdot ||W_V|| \cdot ||v^*||}$
\end{enumerate}
\end{lemma}\label{lemma:whiten}
Detailed proof of all lemmas can be found in Appendix \ref{sec:proof}. 
\begin{remark}
    Lemma 4.9 suggests that when adding new knowledge, a new key must be introduced at an appropriate position. This new key must be placed carefully, as its position can affect both its intended target and potentially interfere with nearby keys.
\end{remark}

\begin{figure}
  \includegraphics[width=\linewidth]{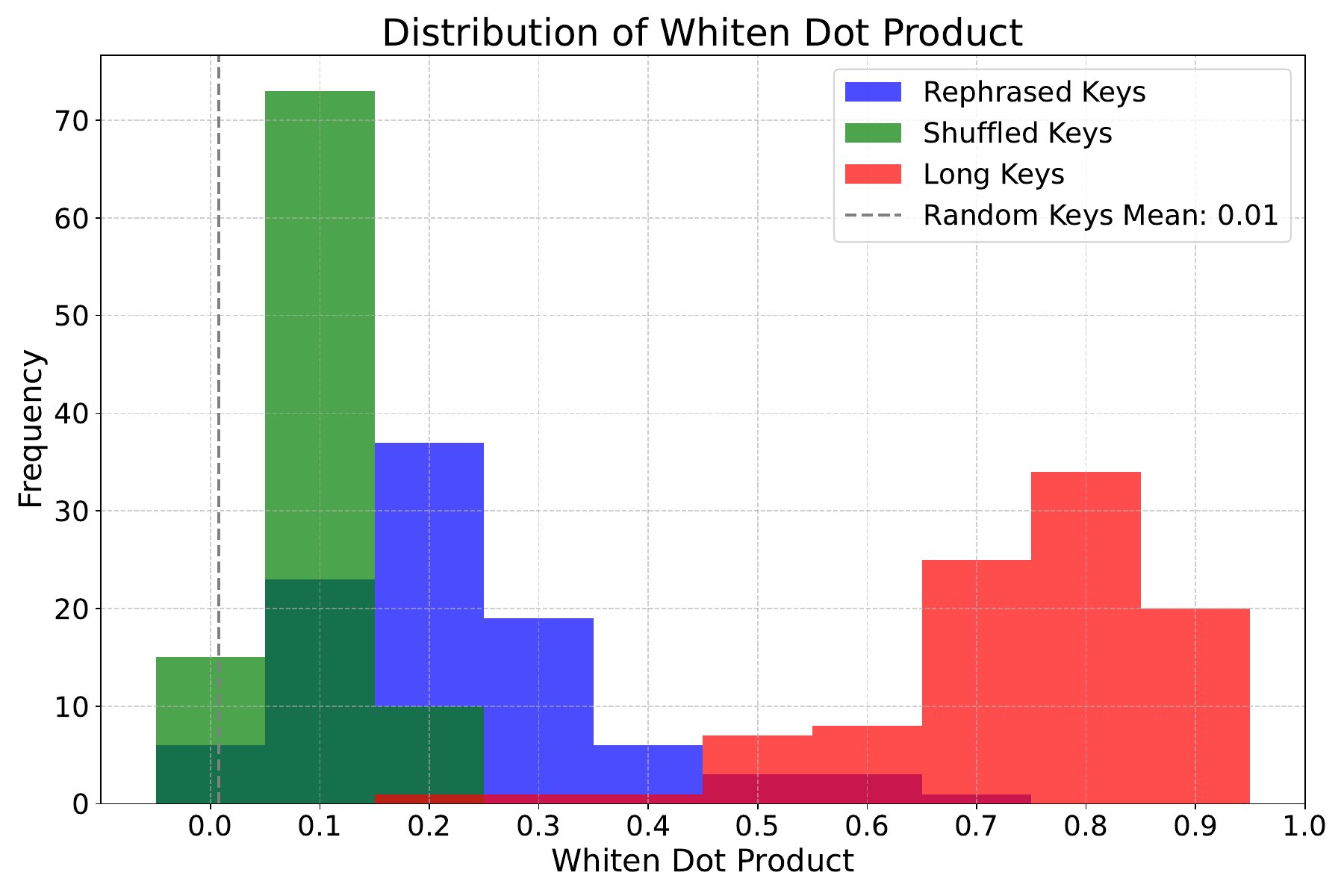}
  \vspace{-20pt}
  \caption{The distribution of normalized whitening similarity between different kinds of keys and original keys.}
  \label{fig:dist_keys}
  \vspace{-20pt}
\end{figure}

\section{Empirical Analysis: A Break of Requirements}
In light of our theoretical results in previous section, we analyze the current knowledge editing methods, showing that the robustness and specificity requirements from previous section cannot be satisfied with inner representations as keys, motivating our approach. 

\subsection{Experimental Setup}
Following previous work, we use the CounterFact~\citep{meng2022locating} datasets, choosing LLaMA-2 as our base model.
In addition to the prompt from CounterFact dataset, we additionally consider three types of perturbation in our experiments, namely prompt appended with unrelated long context, subject rephrase and random shuffled subject. 
Even though the shuffled subject does not contain the same semantic meaning, it demonstrates how keys shift when the position of same token occurs at different positions. 

We collect 10 rephrases for each subject by prompting \texttt{gpt-4o-mini}. The prompt we use can be found in Appendix. 
For long context, we follow \cite{ma2024possible} and extract random text span of 512 tokens from Wikitext-103~\cite{merity2016pointer}. 
For rephrased prompts, we use the paraphrases of prompts released by \cite{patil2023can}. 
For shuffled subject, we sample 10 random orderings of tokens in the subject.
We use 100 samples in our valid set for empirical analyses.

\subsection{Empirical Statistics of Keys, Values and Others}
\label{sec:empirical}

\paragraph{Dissimilar Keys.} \label{section:dissimlar}
In Figure \ref{fig:dist_keys}, we present the distribution of whiten similarity $\beta$ for three operations over the original edit along with a random key baseline. 
The implementation detail can be found in Appendix \ref{appendix:dissimlar}.

We can see that the similarity after these operations drops drastically. 
Rephrasing and shuffling word orders generally reduce the similarity from 1.0 to less than 0.4, even to the random level. Appending long context is less destructive, but still reduces the key similarity to [0.2, 0.9]. 
These results indicate a violation of the robustness requirement, showing a significant variability in the representation of the same subject, making locate-and-edit difficult to retrieve the edited value to be retrieved. 

These findings challenges the intuition that semantically equivalent subjects should have similar representations, and poses severe challenges to the effectiveness of edits. 

\begin{figure*}[t]
    \begin{minipage}[t]{0.45\linewidth}
      \vspace{0pt}  
      \centering
      \includegraphics[width=\linewidth]{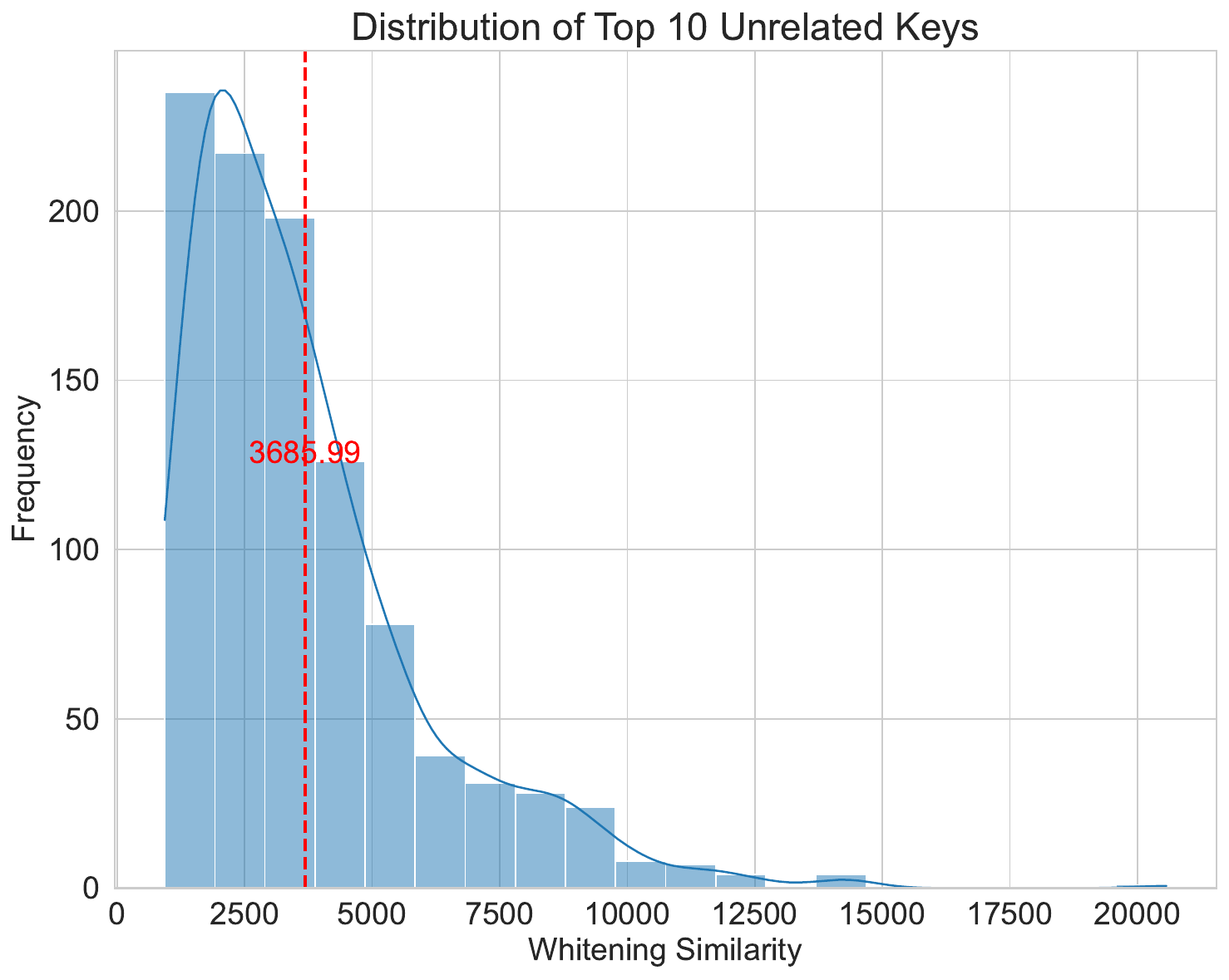}
      \label{fig:your_label}
    \end{minipage}
    \hfill
    \begin{minipage}[t]{0.45\linewidth}
      \vspace{0pt}  
      \centering
      \small
      \resizebox{\textwidth}{!}{
        \begin{tabular}{|c|c|p{3cm}|}
          \hline
          \textbf{Subject} & \textbf{Score} & \textbf{Prefix} \\
          \hline
          \makecell[c]{\itshape Michael Jordan} & \makecell[l]{3898.8} & \makecell[l]{... reach 10,000 career\\ assists. \underline{Kobe}}\\\hline
          \makecell[c]{\itshape Emmitt Smith} & \makecell[l]{8469.9} & \makecell[l]{... for the NL lead with\\Randy Johnson, Kevin\\Millwood, Tom \underline{Glavine}}\\\hline
          \makecell[c]{\itshape Pasquale Di \\\itshape  Sabatino} & \makecell[c]{2190.6} & \makecell[l]{... archbishop of Albi \\ Giovanni Costanzio\\ \underline{Caracciolo}} \\ \hline
          \makecell[c]{\itshape East China \\\itshape Normal University} & \makecell[c]{6798.2} & \makecell[l]{... Located southwest of\\ Gongyi city in Gongxian\\\underline{County}}\\\hline
          \makecell[c]{\itshape BMW Z3} & \makecell[l]{3570.2} & \makecell[l]{... her lead over Röhrl\\shrank to 18 minutes.\\The Toyota \underline{Celica}}\\\hline
        \end{tabular}}
    \end{minipage}
    \vspace{-15pt}
  \caption{\textbf{Left}: CounterFact subjects have unrelated \textcolor[rgb]{1,0,0}{prefixes} which are close in keys. {The red dashed line indicates random keys baseline.} \textbf{Right}: Semantically similar subjects bring challenges to specificity. }
  \vspace{-15pt}
  \label{fig:unrelated}
  \end{figure*}

\paragraph{Similar but Non-Related Keys.}
We also investigate whether there exist different subjects that have highly similar keys. 
To this end, we iterate through a slice of Wikitext-103 dataset~(about 80M tokens) and select those close to subjects in CounterFact in the whitening space. 
We filter those tokens whose prefix has the same subject token and collect the top 10 unrelated keys of each subject in CounterFact. 
The left of Figure \ref{fig:unrelated} plots the distribution of whitening similarities between unrelated prefixes and CounterFact subjects. We find that a large portion of them has an extremely high whitening similarity score, i.e., > 2500. 
Based on our theory, it indicates that any edit that affects these subjects would inevitably affect the output on these unrelated prefixes. 

On the right side of Figure \ref{fig:unrelated}, we present a list of subjects and their top-1 prefix in terms of whitening similarities. 
Interestingly, we observe that a subject can exhibit similarity in distributional semantics~\citep{lenci2023distributional} to its corresponding top unrelated prefix. 
For example, the keys of \textit{Michael Jordan} are highly similar to keys of a prefix related to \emph{Kobe}. 
Considering that these two basketball players has much in common in many perspectives, it makes sense that their keys are similar. 
However, an edit to \emph{Michael Jordan} affects \emph{Kobe} would be definitely unreasonable. 




\section{Robust Edit Pathway}
Our solution is to separate keys from the model's internal representations by introducing a potential branching path as keys for edited facts.

This is done by adding an adapter after the keys, allowing their representations to be modified when needed.
As shown on the right of Figure \ref{fig:overall}, our adapter consists of two modules, a \textit{projection} module that is responsible for aligning the keys and a \text{gate} module that activates the adapter when a token representation needs to be edited:
\begin{gather}
    \hat{k} = f_{\text{gate}}(k) \circ f_{\text{proj}}(k) + k,
\end{gather}
where $k \in \R^{bsz \times L \times D}$, $f_{\text{gate}}(k) \in \R^{bsz \times L \times 1}$ and $f_{\text{proj}}(k) \in \R^{bsz \times L \times D}$.

The gate mechanism here operates on the granularity of tokens and adaptively selects whether a key should be modified or not.

We train the adapter by aggregating the keys of same subject $k_s \in \mathcal{K}_s$ toward our injected target key $k_*$:
\begin{gather}
    \mathcal{L}_\text{agg} = -{|(\frac{\hat{k_s}}{||\hat{k_s}||_2})^T C^{-1} k_*|}
\end{gather}
where $\hat{k_s}$ is the output keys after adapter. 
The intuition is inspired by Lemma \ref{lemma:generalize} and \ref{lemma:spec}. If the edited key is close to the keys of the same subject, especially those we found dissimilar in Section \ref{sec:empirical}, the edit would be more robust. 

In practice, we find that the model inclines to `cheat' by simply increasing the norm of $k_s$, and thus we normalize the output of $f$. 
In practice, we take the last token of rephrased subjects over different contexts and rephrased templates as $k_s$.
This objective, built on the whiten similarity, further strengthens the validness of our {theoretical} results. 

To address the drift of the target key $k_*$ during optimization, we introduce a target consistency loss:
\begin{gather}
    \mathcal{L}_\text{consistency} = MSE(\hat{k_*}, k_*)
\end{gather}
The final training objective combines both components:
\begin{gather} \label{eq:final_training}
    \mathcal{L} = \mathcal{L}_\text{agg} + \alpha \mathcal{L}_\text{consistency}
\end{gather}
with $\alpha$ controlling the trade-off.

For testing, we use a gate threshold $\tau$ to determine whether to activate this projection. 
This gate mechanism allows the model to dynamically decide whether the original keys should be modified. If not, the keys are left intact and thus ensure the locality of edits. 
The whole algorithm can be found in Appendix. 

\begin{table*}[ht]
\resizebox{\textwidth}{!}{\begin{tabular}{c|cccc|ccc|ccc}
\toprule
\multirow{2}{*}{\textbf{Method}} &
  \multicolumn{4}{c|}{\textbf{Edit Performance}} &
  \multicolumn{3}{c|}{\textbf{In-Domain}} &
  \multicolumn{3}{c}{\textbf{Out-of-Domain}} \\\cline{2-11}
 &
  \textbf{Sucess}$\uparrow$ &
  \textbf{Locality}$\uparrow$ &
  \textbf{Para.}$\uparrow$ &
  \textbf{Fluency}$\downarrow$ &
  \textbf{Rephrase}$\uparrow$ &
  \textbf{Shuffle}$\uparrow$ &
  \textbf{Long}$\uparrow$ &
  \textbf{Rephrase}$\uparrow$ &
  \textbf{Shuffle}$\uparrow$ &
  \textbf{Long}$\uparrow$ \\\hline\hline
\multicolumn{11}{c}{\textit{Baseline Methods}}                                                       \\\hline\midrule
\textbf{ROME}       & 100.0 & 96.1 & 63.8 & 587.4 & 61.0 & 13.0 & 89.8 & 62.6 & 13.7 & 89.8 \\
\textbf{MEMIT}      & 99.3  & 91.2 & 71.9 & 571.4 & 73.3 & 30.0 & 92.3 & 73.4 & 32.0 & 94.3 \\
\textbf{R-ROME}     & 99.7  & 95.8 & 62.1 & 583.8 & 58.9 & 14.7 & 89.5 & 61.7 & 16.1 & 90.7 \\
\textbf{EMMET}      & 99.7  & 93.8 & 63.0 & 584.0 & 59.7 & 16.3 & 83.7 & 60.9 & 16.5 & 83.0 \\\hline\midrule
\multicolumn{11}{c}{\textit{With REP}}                                                               \\\hline\midrule
\textbf{ROME}   & 100.0\textsuperscript{+0.0} & 94.6\textsuperscript{-1.5} & 66.9\textsuperscript{+3.1} & 587.5\textsuperscript{+0.1} & 88.0\textsuperscript{+27.0} & 59.9\textsuperscript{+46.9} & 91.7\textsuperscript{+1.9} & 75.5\textsuperscript{+12.9} & 28.7\textsuperscript{+15.0} & 91.3\textsuperscript{+1.5} \\
\textbf{MEMIT}  & 99.4\textsuperscript{+0.1}  & 90.8\textsuperscript{-0.4} & 74.2\textsuperscript{+2.3} & 567.2\textsuperscript{-4.2} & 89.9\textsuperscript{+16.6} & 58.9\textsuperscript{+28.9} & 93.6\textsuperscript{+1.3} & 84.4\textsuperscript{+11.0} & 45.2\textsuperscript{+31.5} & 94.2\textsuperscript{-0.1} \\
\textbf{R-ROME} & 99.9\textsuperscript{+0.2}  & 94.7\textsuperscript{-1.1} & 67.4\textsuperscript{+5.3} & 586.0\textsuperscript{+2.2} & 88.8\textsuperscript{+29.9} & 60.3\textsuperscript{+45.6} & 92.0\textsuperscript{+2.5} & 76.5\textsuperscript{+14.8} & 29.5\textsuperscript{+13.4} & 92.0\textsuperscript{+1.3} \\
\textbf{EMMET}  & 99.8\textsuperscript{+0.1}  & 92.2\textsuperscript{-1.6} & 68.4\textsuperscript{+5.4} & 584.6\textsuperscript{+0.6} & 94.4\textsuperscript{+34.7} & 82.7\textsuperscript{+66.4} & 88.4\textsuperscript{+4.7} & 82.9\textsuperscript{+22.0} & 42.5\textsuperscript{+26.0} & 88.6\textsuperscript{+5.6} \\\bottomrule
\end{tabular}}
\caption{\textbf{The main results of REP across three seeds comparing ROME, MEMIT, R-ROME, and EMMET editing methods on Llama2-7B on CounterFact dataset}. REP consistently enhances model performance
Results averaged over three seeds with $\tau = 0.9$. The upperscript numbers denote the improvement after using REP. 
}
\vspace{-18pt}
\label{tab:llama2}
\end{table*}

\subsection{Experimental Results}

\paragraph{Setup}
We evaluate our Robust Edit Pathway with representative locate-and-edit methods, namely ROME, MEMIT, R-ROME, and EMMET. 
We use the LLaMA2-7B, Mistral-7B, and Qwen2-7B as our base model and CounterFact and ZsRE as our datasets.
We filter knowledge triplets of datasets not presented in the model as \cite{meng2022locating} did, and randomly sample 100 knowledge triplets as the validation set and 400 triplets as the test set. 
While other studies in model editing explore modifying multiple facts continuously~\citep{mitchell2022memory,hartvigsen2024aging,meng2022mass}, we have found that robustly injecting even a single fact presents significant challenges. Therefore,  we keep our focus on single-edit paradigm.

For evaluation, we first follow \citep{meng2022locating,meng2022mass} and utilize the following four metrics for edit performance: (1) \emph{Success}: the ratio of targeted knowledge achieving the top probability; (2) \emph{Locality}: the {ratio} of related but non-identical facts kept intact by the edit; 
(3) Paraphrase (Para.): the ratio of targeted knowledge achieving success on paraphrased prompts;
(4) Fluency: the weighted average of bi- and tri-gram entropies. 

Moreover, we report the success rate for three robustness tests: paraphrasing subjects, shuffling subjects' token ordering, and appending long context, as discussed throughout the paper. 
Improving these metrics suggests a more robust editing method. 
We report robustness metrics at both in-domain, where the test cases are seen in training adapter, and out-of-domain, where the test cases are not seen by adapter. 
Note that in our `in-domain', we do not reveal the target knowledge to the model, we only aggregate the keys. 



\paragraph{ROME and MEMIT's Failure on Robustness. }
Our results are shown in Table \ref{tab:llama2}. 
We can see that our baseline methods, ROME, MEMIT, R-ROME, and EMMET achieve near-perfect edit success rates~(>99\%) while preserving good locality scores~(93-96\%). 
Nonetheless, these methods are prune to robustness tests. 
Taking ROME as an example, the success rate drops 39\% with rephrased subjects, 87\% with shuffled subjects ordering, and 10.2\% with randomly appended long context. 
These results reconcile with those reported in previous studies~\citep{ma2024possible}.

\paragraph{Effectiveness of Robust Edit Pathway.} 
REP improves the robustness of each of the \emph{locate-and-edit} methods significantly, with a slight cost of locality drop. For instance, REP improves ROME over three robustness tests with +27.0\%/+46.9\%/+1.9\% for in-domain queries, and +12.9\%/+15.0\%/+1.5\% for out-of-domain queries. 
We also conduct experiments over a different dataset~(ZsRE) and two additional base models~(QWen and Mistral). The results are shown in Appendix and consistently demonstrate the effectiveness of REP. 
This validates our theoretical results and empirical insights.

\paragraph{Ablation Studies.} \label{sec:ablation}

Gate threshold $\tau$ and consistency loss weight $\alpha$ are crucial to the performance of REP. 
We study them in the Appendix with Figure \ref{ablation:tau} and Figure \ref{ablation:target_loss}.
We find that a larger $\tau$ and a larger $\alpha$ leads to a better locality and success rate. Meanwhile, the robustness metrics first plateau, then degrade with the increase of $\tau$ and $\alpha$, indicating a trade-off between robustness and edit performance. Throughout our experiments, we use $\tau=0.9$ and $\alpha=5e+4$. 








\section{Conclusion}
In this work, we challenge a core assumption in the locate-and-edit mechanism -- \textit{the model's inner representations can serve as semantic keys for editing}. 
We present theoretical results and empirical analyses revealing that these keys are both sensitive and unspecific. To address this issue, we propose the Robust Edit Pathway (REP), which disentangles the editing keys from native model representations. By extensive experiments, we show that REP can significantly enhance robustness over various locate-and-edit methods while maintains the edit success rate. 

\section{Limitations}
While REP demonstrates significant improvements in knowledge editing robustness, our work is limited in the following aspects: 
(1) REP requires additional training steps to learn the adapter parameters, introducing computational overhead compared to direct editing methods. 
(2) Our current evaluation focuses on single-fact editing. The effectiveness of REP in scenarios involving multiple interrelated facts or continuous editing remains to be investigated.
(3) In this work, we focus on locate-and-edit methods. Even though it is the dominant line of methods in model editing methods, there are still other model editing methods and REP does not apply to them. 

\section{Acknowledgement}
This publication has emanated from research conducted with the financial support of the National Natural Science Foundation of China Key Program under Grant Number 6233000066

\bibliography{sources/reference}

\appendix

\clearpage
\appendix

\section{Proofs}
\label{sec:proof}
\subsection{Proof of Lemma \ref{lemma:fuzzy}}
Since $K$ has full row rank ($\text{rank}(K) = D_1$), $K K^T$ is invertible. To find $\alpha$, we use the Moore-Penrose pseudoinverse of $K$.

Given $K \in \R^{D_1\times n}$, the pseudoinverse $K^{+}$ is defined as: $K^{+}=K^T (K K^T)^{-1}$, which also minimizes $||K\alpha - \hat{k}||$. 

Then, we can express $\hat{v}$ as:
\begin{equation}
    \hat{v} = V \alpha = V K^T (K K^T)^{-1} \hat{k}.
\end{equation}

Note that since $n \ggg D_1$, the system $K\alpha = \hat{k}$ is underdetermined. 
This means there are infinitely many solutions for $\alpha$, and the Moore-Penrose pseudoinverse gives the one with the smallest norm.

\subsection{Proof of Lemma \ref{lemma:value_bound}}

We can focus on the logit difference between the largest and the second-largest logits to achieve high confidence in the final prediction. This difference is an important factor in determining the confidence of a prediction in a softmax layer.

Here, we simplify the modeling by only considering the contribution of edited layer towards final prediction via its 
the edited layer is connected to the final prediction layer directly via its attention layer

Given a vector of logits \( \mathbf{z} = [z_1, z_2, \ldots, z_n] \), the softmax function yields probabilities \( \mathbf{p} = [p_1, p_2, \ldots, p_n] \), where:

\[
p_i = \frac{e^{z_i}}{\sum_{j=1}^{n} e^{z_j}}
\]

To increase the confidence in the prediction for the largest logit, maximize the difference between the largest logit and the second-largest logit.

Let \( z_{\text{max}} \) be the largest logit and \( z_{\text{other}} \) be another logit.
The logit difference \( \Delta \) is given by: $\epsilon = z_{\text{max}} - z_{\text{other}}$.

The softmax confidence for the class corresponding to \( z_{\text{max}} \) can be expressed as:

\begin{align}
    p_{\text{max}} &= \frac{e^{z_{\text{max}}}}{e^{z_{\text{max}}} + e^{z_{\text{other}}} + \sum_{k \neq \text{max, other}} e^{z_k}} \\
    & < \frac{e^{z_{\text{max}}}}{e^{z_{\text{max}}} + e^{z_{\text{other}}}} \\
    & = \frac{1}{1 + e^{-\epsilon}}
\end{align}
After organizing between two sides, we get a lower bound of $\epsilon$ for achieving a sufficiently large confidence:
\begin{equation}
\label{eq:softmax_diff}
    \epsilon > - \log (1-\frac{1}{p_{\text{max}}})
\end{equation}

Now, in a transformer architecture, the edited MLP layer is connected to the word prediction layer through an attention layer at the final token. 
Let the difference between the original and the edited output of the MLP layer be $\Delta v$, the parameters of the attention layer are $W_Q, W_K, W_V \in \mathbb{R}^{D \times D}$ and the query vector at the prediction token is $q=Q h$, the attention layer's output is defined by
\begin{align}
    o = \sum_j \text{Softmax}(q^T W_K v_j) W_V v_j.
\end{align}
Since in the locating part we use causal intervention to identify the most influential position of tokens to edit, we can assume that $(q^TW_K v_s)$ has already get the largest weight. 
The difference caused by edited MLP is, 
\begin{align}
    \Delta o = \text{Softmax}(\cdot) W_V \Delta v.
\end{align}
Then, residual connections directly connect this output to the final word prediction layer.
Combining our result from equation \ref{eq:softmax_diff}, let the original fact $t$ before the edit has a logit gap $\epsilon_1$ and the new fact $t_*$ after edit has $\epsilon_2$, we can bound the $\Delta o$ with,
\begin{align}
    &\begin{cases}
    (w_t - w_{t_*})^T o_{ori} > \epsilon_1 \\
    (w_{t_*} - w_t)^T (o_{ori} + \Delta o) > \epsilon_2
    \end{cases} \\
    &\Rightarrow (w_{t_*} - w_t)^T \Delta o > \epsilon_1 + \epsilon_2 \\
    &\Rightarrow (w_{t_*} - w_t)^T \text{Softmax}(\cdot) W_V \Delta v > \epsilon_1 + \epsilon_2 \\
    &\Rightarrow (w_{t_*} - w_t)^T W_V \Delta v > \epsilon_1 + \epsilon_2  \\
\end{align}
Given that the softmax weight is at most 1, we have our lower bound on $\Delta v$. 

\subsection{Proof of Lemma \ref{lemma:generalize}}
For an edit to be robust, it must propagate correctly to all semantically equivalent inputs. We derive this requirement step by step:

1) From Lemma 4.4, a successful edit requires:
\begin{gather*}
    (w_{t^*} - w_t)^TW_V(v^* - v_o) > \epsilon_1 + \epsilon_2
\end{gather*}

2) When querying with a semantically equivalent key $k_s \in K_s$, by Lemma 4.1, the retrieved value is:
\begin{gather*}
    v = k_s^TC^{-1}k^* \cdot v^* = \beta_{s,*} \cdot v^*
\end{gather*}
where $\beta_{s,*}$ represents the whiten similarity between $k_s$ and $k^*$.

3) For robust editing, this retrieved value must maintain the prediction gap:
\begin{gather*}
    (w_{t^*} - w_t)^TW_V(\beta_{s,*} \cdot v^*) > \epsilon_1 + \epsilon_2
\end{gather*}

4) Rearranging terms:
\begin{gather*}
    (w_{t^*} - w_t)^TW_V(k_s^TC^{-1}k^*) \cdot v^{*T} > \epsilon_1 + \epsilon_2, \quad \\
    \forall k_s \in K_s
\end{gather*}

This inequality must hold for all semantically equivalent keys $k_s \in K_s$, establishing our robustness requirement.

\subsection{Proof of Lemma \ref{lemma:spec}}
The specificity requirement ensures edits do not affect unrelated knowledge. We derive this as follows:

1) Consider an unrelated key $k_o \notin K_s$ with original target $t_n$. The corresponding output embedding is $w_n$.

2) To preserve specificity, the edit should not significantly alter predictions for unrelated inputs:
\begin{gather*}
    (w_n - w_{t^*})^TW_V(k_o^TC^{-1}k^*) \cdot v^{*T} < \epsilon_3
\end{gather*}

3) This constraint must hold for:
\begin{itemize}
    \item All unrelated keys $k_o \notin K_s$
    \item All possible target embeddings $w_n \in W$
\end{itemize}

4) Therefore, our specificity requirement is:
\begin{gather*}
    (w_n - w_{t^*})^TW_V(k_o^TC^{-1}k^*) \cdot v^{*T} < \epsilon_3, \\ \forall k_o \notin K_s, \forall w_n \in W
\end{gather*}

This establishes the formal criterion for maintaining specificity in knowledge editing. The requirement ensures that edits remain localized to the intended knowledge while not affecting unrelated retrievals.









\section{Experimental Details}

\subsection{Data Construction}
We build our evaluation data based on the CounterFact dataset. We further augment our data with all three robustness tests. 
For rephrased subjects by prompting \texttt{gpt4o-mini} with the following prompt.
\begin{quote}
    Give 10 rephrases representing the same entity: \{ENTITY\}
\end{quote}
The irrelevant long contexts are extracted from the Wikitext-103 dataset~\citep{merity2016pointer}. The shuffled tokens are generated via sampling different word ordering. 
Finally, we filter the samples that are not present in the current LLM, that is, given the prefix, the target tokens are not predicted by the LLMs with the top-1 probabilities. 
We sample 100 samples for validation and 400 samples for test. 
To evaluate in-domain and out-of-domain robustness, we split the all three kinds of robustness queries in a 50-50 manner. 
For each sample, we have 5 in-domain queries and 5 out-of-domain queries. 

\subsection{Analyzing Dissimilar Keys} \label{appendix:dissimlar}
In Section \ref{section:dissimlar}, for each subject in CounterFact, we compute the dot product for each pair of keys of a subject's rephrases. 
We utilize the inputs to the FFN's down projection of layer 5 of LLaMA-2 as our keys, consistent with previous ROME experiments. 
Additionally, we include the dot product values of randomly sampled keys as a baseline for comparative analysis. 
We normalize the whiten similarity by the similarities between the subject itself.

\subsection{Details of Training REP}
We implement our methods based on EasyEdit~\citep{wang2023easyedit}.
We use Adam optimizer for all experiments and the learning rate is 5e-4.
We train each adaptor for 10 steps. 
The inner dimension of the projection module is 32, and the inner dimension of gate module is 0.1 of key dimension. 


\section{Additional Results}
Table \ref{tab:main} and \ref{tab:zsre} present the performance of ROME, MEMIT, R-ROME, and EMMET methods, both with and without the REP enhancement, across CounterFact and ZSRE respectively.
Across both datasets, REP consistently improves model robustness, particularly in in-domain generalization and out-of-domain adaptability, despite minor trade-offs in edit success rates. Results are averaged over three seeds , with standard deviations indicating stable improvements. Notably, REP-enhanced variants demonstrate superior fluency and locality preservation, highlighting its effectiveness in balancing edit precision with broader generalization.

\begin{table*}[h]
    \caption{\textbf{The main results of REP across three seeds comparing ROME, MEMIT, R-ROME, and EMMET editing methods on Llama2-7B, Mistral-7B and Qwen2-7b on CounterFact dataset}. REP consistently enhances model performance
    Results averaged over three seeds with $\tau = 0.9$, showing standard deviations. $\uparrow$ indicates higher values are better, $\downarrow$ indicates lower values are better.}\label{tab:main}
    \resizebox{\textwidth}{!}{%
    \begin{tabular}{ccccccccccccc}
    \toprule
    \multicolumn{1}{l}{}                             & \multicolumn{1}{l}{}                     & \multicolumn{3}{c}{\textbf{Edit Performance}}                                       & \multicolumn{2}{c}{\textbf{Generalization}}                     & \multicolumn{3}{c}{\textbf{In-Domain}}                                               & \multicolumn{3}{c}{\textbf{Out-of-Domain}}                      \\ \midrule
    \multicolumn{1}{c|}{\textit{Model}}                       & \multicolumn{1}{c|}{\textit{Method}}     & \textit{Success}$\uparrow$     & \textit{Locality}$\uparrow$   & \multicolumn{1}{c|}{\textit{Reversion}$\downarrow$}& \textit{Para.}$\uparrow$        & \multicolumn{1}{c|}{\textit{Fluency}$\uparrow$ }    & \textit{Rephrase}$\uparrow$   & \textit{Shuffle}$\uparrow$    & \multicolumn{1}{c|}{\textit{Long}$\uparrow$}       & \textit{Rephrase}$\uparrow$   & \textit{Shuffle}$\uparrow$    & \textit{Long}$\uparrow$       \\ \midrule
    \multicolumn{1}{c|}{\multirow{8}{*}{\texttt{Llama2}}}  & \multicolumn{1}{c|}{\textbf{ROME}}       & 100.0 ± 0.0         & \textbf{96.1 ± 0.1} & \multicolumn{1}{c|}{0.0 ± 0.0}          & 63.8 ± 0.3         & \multicolumn{1}{c|}{587.4 ± 1.2}          & 61.0 ± 0.7          & 13.0 ± 0.9          & \multicolumn{1}{c|}{89.8 ± 0.2}          & 62.6 ± 0.1          & 13.7 ± 0.5          & 89.8 ± 0.5          \\
    \multicolumn{1}{c|}{}                            & \multicolumn{1}{c|}{\textbf{ +REP}}   & 100.0 ± 0.0         & 94.6 ± 0.2          & \multicolumn{1}{c|}{0.0 ± 0.0}          & \textbf{66.9 ± 0.3} & \multicolumn{1}{c|}{\textbf{587.5 ± 0.8}} & \textbf{88.0 ± 0.2} & \textbf{59.9 ± 0.3} & \multicolumn{1}{c|}{\textbf{91.7 ± 0.2}} & \textbf{75.5 ± 0.6} & \textbf{28.7 ± 1.9} & \textbf{91.3 ± 1.4} \\
    \multicolumn{1}{c|}{}                            & \multicolumn{1}{c|}{\textbf{MEMIT}}      & 99.3 ± 0.5          & \textbf{91.2 ± 0.6} & \multicolumn{1}{c|}{0.0 ± 0.0}          & 71.9 ± 1.7          & \multicolumn{1}{c|}{\textbf{571.4 ± 2.6}} & 73.3 ± 1.2          & 30.0 ± 0.9          & \multicolumn{1}{c|}{92.3 ± 0.9}          & 73.4 ± 0.7          & 32.0 ± 3.1          & \textbf{94.3 ± 3.3} \\
    \multicolumn{1}{c|}{}                            & \multicolumn{1}{c|}{\textbf{+REP}}  & \textbf{99.4 ± 0.1} & 90.8 ± 0.2          & \multicolumn{1}{c|}{0.0 ± 0.0}          & \textbf{74.2 ± 0.1} & \multicolumn{1}{c|}{567.2 ± 0.3}          & \textbf{89.9 ± 0.4} & \textbf{58.9 ± 0.8} & \multicolumn{1}{c|}{\textbf{93.6 ± 1.1}} & \textbf{84.4 ± 0.5} & \textbf{45.2 ± 0.8} & 94.2 ± 1.5          \\
    \multicolumn{1}{c|}{}                            & \multicolumn{1}{c|}{\textbf{R-ROME}}     & 99.7 ± 0.5          & \textbf{95.8 ± 0.3} & \multicolumn{1}{c|}{0.3 ± 0.5}          & 62.1 ± 1.3          & \multicolumn{1}{c|}{583.8 ± 3.3}          & 58.9 ± 0.7          & 14.7 ± 0.8          & \multicolumn{1}{c|}{89.5 ± 3.5}          & 61.7 ± 1.3          & 16.1 ± 1.8          & 90.7 ± 0.5          \\
    \multicolumn{1}{c|}{}                            & \multicolumn{1}{c|}{\textbf{+REP}} & \textbf{99.9 ± 0.1} & 94.7 ± 0.4          & \multicolumn{1}{c|}{\textbf{0.0 ± 0.0}} & \textbf{67.4 ± 0.2} & \multicolumn{1}{c|}{\textbf{586.0 ± 0.1}} & \textbf{88.8 ± 0.5} & \textbf{60.3 ± 1.2} & \multicolumn{1}{c|}{\textbf{92.0 ± 0.2}} & \textbf{76.5 ± 0.6} & \textbf{29.5 ± 1.4} & \textbf{92.0 ± 0.8} \\
    \multicolumn{1}{c|}{}                            & \multicolumn{1}{c|}{\textbf{EMMET}}      & 99.7 ± 0.5          & \textbf{93.8 ± 0.2} & \multicolumn{1}{c|}{\textbf{0.0 ± 0.0}} & 63.0 ± 1.3          & \multicolumn{1}{c|}{584.0 ± 6.5}          & 59.7 ± 2.3          & 16.3 ± 0.6          & \multicolumn{1}{c|}{83.7 ± 0.2}          & 60.9 ± 1.2          & 16.5 ± 1.5          & 83.0 ± 2.4          \\
    \multicolumn{1}{c|}{}                            & \multicolumn{1}{c|}{\textbf{+REP}}  & \textbf{99.8 ± 0.2} & 92.2 ± 0.4          & \multicolumn{1}{c|}{0.1 ± 0.1}          & \textbf{68.4 ± 0.2} & \multicolumn{1}{c|}{\textbf{584.6 ± 0.8}} & \textbf{94.4 ± 0.2} & \textbf{82.7 ± 1.0} & \multicolumn{1}{c|}{\textbf{88.4 ± 1.9}} & \textbf{82.9 ± 0.3} & \textbf{42.5 ± 1.3} & \textbf{88.6 ± 2.2} \\ \midrule
    \multicolumn{1}{c|}{\multirow{8}{*}{\texttt{Mistral}}} & \multicolumn{1}{c|}{\textbf{ROME}}       & \textbf{99.9 ± 0.1} & \textbf{94.1 ± 0.0} & \multicolumn{1}{c|}{\textbf{0.0 ± 0.0}} & 69.1 ± 0.5          & \multicolumn{1}{c|}{609.4 ± 0.8}          & 71.1 ± 0.2          & 14.6 ± 0.2          & \multicolumn{1}{c|}{94.6 ± 0.4}          & 71.8 ± 0.3          & 14.3 ± 1.1          & 94.4 ± 0.3          \\
    \multicolumn{1}{c|}{}                            & \multicolumn{1}{c|}{\textbf{+REP}}   & 99.8 ± 0.2          & 92.8 ± 0.1          & \multicolumn{1}{c|}{0.1 ± 0.1}          & \textbf{72.2 ± 0.2} & \multicolumn{1}{c|}{\textbf{610.0 ± 0.5}} & \textbf{95.5 ± 0.2} & \textbf{84.6 ± 0.6} & \multicolumn{1}{c|}{\textbf{95.1 ± 0.4}} & \textbf{84.8 ± 0.8} & \textbf{41.6 ± 0.5} & \textbf{94.7 ± 0.3} \\
    \multicolumn{1}{c|}{}                            & \multicolumn{1}{c|}{\textbf{MEMIT}}      & \textbf{99.7 ± 0.3} & \textbf{89.2 ± 0.2} & \multicolumn{1}{c|}{0.0 ± 0.0}          & 76.8 ± 0.5          & \multicolumn{1}{c|}{\textbf{607.0 ± 0.9}} & 84.0 ± 0.1          & 29.0 ± 0.3          & \multicolumn{1}{c|}{95.0 ± 0.6}          & 82.4 ± 0.4          & 28.6 ± 0.4          & 94.0 ± 0.5          \\
    \multicolumn{1}{c|}{}                            & \multicolumn{1}{c|}{\textbf{+REP}}  & 98.5 ± 0.5          & 85.5 ± 0.1          & \multicolumn{1}{c|}{0.0 ± 0.0}          & \textbf{77.3 ± 0.6} & \multicolumn{1}{c|}{605.5 ± 1.0}          & \textbf{93.1 ± 0.5} & \textbf{75.1 ± 1.3} & \multicolumn{1}{c|}{\textbf{95.4 ± 0.4}} & \textbf{89.2 ± 0.3} & \textbf{62.7 ± 0.7} & \textbf{94.3 ± 0.3} \\
    \multicolumn{1}{c|}{}                            & \multicolumn{1}{c|}{\textbf{R-ROME}}     & \textbf{99.8 ± 0.1} & \textbf{93.7 ± 0.1} & \multicolumn{1}{c|}{\textbf{0.0 ± 0.0}} & 70.5 ± 0.1          & \multicolumn{1}{c|}{608.6 ± 0.9}          & 73.2 ± 0.2          & 16.1 ± 0.4          & \multicolumn{1}{c|}{95.5 ± 0.4}          & 73.4 ± 0.3          & 16.0 ± 1.2          & 95.3 ± 1.2          \\
    \multicolumn{1}{c|}{}                            & \multicolumn{1}{c|}{\textbf{+REP}} & 99.7 ± 0.1          & 92.4 ± 0.0          & \multicolumn{1}{c|}{0.1 ± 0.1}          & \textbf{73.6 ± 0.1} & \multicolumn{1}{c|}{\textbf{609.4 ± 1.2}} & \textbf{96.1 ± 0.1} & \textbf{86.9 ± 0.3} & \multicolumn{1}{c|}{\textbf{95.9 ± 0.2}} & \textbf{85.9 ± 0.2} & \textbf{43.9 ± 0.5} & \textbf{95.4 ± 1.0} \\
    \multicolumn{1}{c|}{}                            & \multicolumn{1}{c|}{\textbf{EMMET}}      & \textbf{99.8 ± 0.1} & \textbf{92.5 ± 0.2} & \multicolumn{1}{c|}{0.1 ± 0.1}          & 69.6 ± 0.9          & \multicolumn{1}{c|}{\textbf{609.0 ± 1.0}} & 73.8 ± 0.5          & 16.7 ± 0.5          & \multicolumn{1}{c|}{92.0 ± 0.7}          & 73.5 ± 0.3          & 16.4 ± 0.6          & 91.4 ± 1.3          \\
    \multicolumn{1}{c|}{}                            & \multicolumn{1}{c|}{\textbf{+REP}}  & 99.0 ± 0.0          & 89.9 ± 0.3          & \multicolumn{1}{c|}{0.1 ± 0.1}          & \textbf{74.2 ± 1.1} & \multicolumn{1}{c|}{608.4 ± 0.3}          & \textbf{98.2 ± 0.1} & \textbf{95.2 ± 1.0} & \multicolumn{1}{c|}{\textbf{93.6 ± 0.7}} & \textbf{90.2 ± 0.3} & \textbf{58.7 ± 1.0} & \textbf{92.9 ± 1.5} \\ \midrule
    \multicolumn{1}{c|}{\multirow{8}{*}{\texttt{Qwen2}}}   & \multicolumn{1}{c|}{\textbf{ROME}}       & \textbf{99.6 ± 0.1} & \textbf{95.6 ± 0.1} & \multicolumn{1}{c|}{0.0 ± 0.0}          & 69.4 ± 0.3          & \multicolumn{1}{c|}{620.4 ± 1.5}          & 63.2 ± 0.3          & 20.0 ± 0.4          & \multicolumn{1}{c|}{94.1 ± 0.3}          & 62.7 ± 0.2          & 18.1 ± 0.5          & 93.9 ± 0.5          \\
    \multicolumn{1}{c|}{}                            & \multicolumn{1}{c|}{\textbf{+REP}}   & 99.4 ± 0.1          & 91.0 ± 0.1          & \multicolumn{1}{c|}{0.0 ± 0.0}          & \textbf{73.1 ± 0.1} & \multicolumn{1}{c|}{\textbf{622.1 ± 1.9}} & \textbf{81.0 ± 0.6} & \textbf{70.5 ± 0.5} & \multicolumn{1}{c|}{\textbf{95.9 ± 0.2}} & \textbf{75.6 ± 0.0} & \textbf{65.4 ± 0.6} & \textbf{95.8 ± 0.5} \\
    \multicolumn{1}{c|}{}                            & \multicolumn{1}{c|}{\textbf{MEMIT}}      & 99.6 ± 0.1          & \textbf{90.3 ± 0.2} & \multicolumn{1}{c|}{0.4 ± 0.1}          & 75.6 ± 0.1          & \multicolumn{1}{c|}{620.1 ± 0.3}          & 75.9 ± 0.6          & 31.4 ± 0.9          & \multicolumn{1}{c|}{97.6 ± 0.1}          & 74.8 ± 0.4          & 29.7 ± 0.9          & 96.6 ± 0.3          \\
    \multicolumn{1}{c|}{}                            & \multicolumn{1}{c|}{\textbf{+REP}}  & \textbf{99.7 ± 0.1} & 81.8 ± 0.1          & \multicolumn{1}{c|}{\textbf{0.0 ± 0.0}} & \textbf{79.7 ± 0.2} & \multicolumn{1}{c|}{\textbf{620.2 ± 2.2}} & \textbf{95.7 ± 0.2} & \textbf{81.7 ± 1.2} & \multicolumn{1}{c|}{\textbf{98.0 ± 0.1}} & \textbf{89.7 ± 0.1} & \textbf{72.3 ± 1.5} & \textbf{96.9 ± 0.2} \\
    \multicolumn{1}{c|}{}                            & \multicolumn{1}{c|}{\textbf{R-ROME}}     & 99.8 ± 0.0          & \textbf{96.2 ± 0.1} & \multicolumn{1}{c|}{0.2 ± 0.0}          & 68.5 ± 0.4          & \multicolumn{1}{c|}{621.0 ± 0.4}          & 63.1 ± 0.3          & 20.2 ± 0.3          & \multicolumn{1}{c|}{93.8 ± 0.2}          & 62.3 ± 0.1          & 18.4 ± 0.4          & 93.3 ± 0.9          \\
    \multicolumn{1}{c|}{}                            & \multicolumn{1}{c|}{\textbf{+REP}} & \textbf{99.9 ± 0.1} & 91.9 ± 0.1          & \multicolumn{1}{c|}{\textbf{0.1 ± 0.1}} & \textbf{72.4 ± 0.5} & \multicolumn{1}{c|}{\textbf{621.1 ± 0.2}} & \textbf{81.4 ± 0.3} & \textbf{70.8 ± 0.8} & \multicolumn{1}{c|}{\textbf{95.7 ± 0.5}} & \textbf{75.5 ± 0.2} & \textbf{64.8 ± 0.9} & \textbf{94.5 ± 1.1} \\
    \multicolumn{1}{c|}{}                            & \multicolumn{1}{c|}{\textbf{EMMET}}      & 99.8 ± 0.0          & \textbf{92.5 ± 0.1} & \multicolumn{1}{c|}{0.2 ± 0.0}          & 72.2 ± 0.1          & \multicolumn{1}{c|}{619.5 ± 0.2}          & 71.5 ± 0.7          & 31.3 ± 0.6          & \multicolumn{1}{c|}{96.2 ± 0.4}          & 70.5 ± 0.9          & 30.0 ± 0.4          & 96.7 ± 0.6          \\
    \multicolumn{1}{c|}{}                            & \multicolumn{1}{c|}{\textbf{+REP}}  & \textbf{99.9 ± 0.1} & 76.4 ± 0.8          & \multicolumn{1}{c|}{\textbf{0.0 ± 0.0}} & \textbf{78.5 ± 0.4} & \multicolumn{1}{c|}{\textbf{621.0 ± 2.6}} & \textbf{92.9 ± 0.3} & \textbf{88.4 ± 0.9} & \multicolumn{1}{c|}{\textbf{97.0 ± 0.1}} & \textbf{89.2 ± 0.3} & \textbf{87.1 ± 1.3} & \textbf{97.4 ± 0.6} \\ \bottomrule
    \end{tabular}%
    }
    \vspace{-10pt}
\end{table*}

\begin{table*}[h]
\caption{The main results of REP across three seeds comparing ROME, MEMIT, R-ROME, and EMMET editing methods on Llama2-7B, Mistral-7B and Qwen2-7b on ZSRE dataset. REP consistently enhances model performance. Results averaged over three seeds with $\tau = 0.9$, showing standard deviations. $\uparrow$ indicates higher values are better, $\downarrow$ indicates lower values are better.}
\label{tab:zsre}
\resizebox{\textwidth}{!}{%
\begin{tabular}{ccccc|c|cccccc}
\toprule
\multicolumn{1}{l}{}                          & \multicolumn{1}{l}{}                 & \multicolumn{3}{c|}{\textbf{Edit Performance}}                 & \textbf{Generalization} & \multicolumn{3}{c}{\textbf{In-Domain}}                                               & \multicolumn{3}{c}{\textbf{Out-of-Domain}}                      \\ \midrule
\multicolumn{1}{c|}{\textit{Model}}           & \multicolumn{1}{c|}{\textit{Method}} & \textit{Sucess}     & \textit{Locality}   & \textit{Reversion} & \textit{Fluency}        & \textit{Rephrase}   & \textit{Shuffle}    & \multicolumn{1}{c|}{\textit{Long}}       & \textit{Rephrase}   & \textit{Shuffle}    & \textit{Long}       \\ \midrule
\multicolumn{1}{c|}{\multirow{8}{*}{\texttt{Llama2}}}  & \multicolumn{1}{c|}{\textbf{ROME}}   & \textbf{92.1 ± 0.1} & 99.6 ± 0.0          & \textbf{0.5 ± 0.0} & 566.1 ± 1.8             & 44.4 ± 0.3          & 4.7 ± 0.1           & \multicolumn{1}{c|}{68.2 ± 0.6}          & 44.2 ± 0.9          & 4.5 ± 0.3           & 68.3 ± 1.0          \\
\multicolumn{1}{c|}{}                         & \multicolumn{1}{c|}{\textbf{+REP}}   & 90.0 ± 0.5          & 99.6 ± 0.0          & 0.6 ± 0.1          & \textbf{567.2 ± 1.8}    & \textbf{72.3 ± 0.2} & \textbf{51.5 ± 0.3} & \multicolumn{1}{c|}{\textbf{72.5 ± 0.2}} & \textbf{58.0 ± 0.7} & \textbf{24.9 ± 0.4} & \textbf{71.2 ± 1.7} \\
\multicolumn{1}{c|}{}                         & \multicolumn{1}{c|}{\textbf{MEMIT}}  & \textbf{88.5 ± 0.6} & 99.4 ± 0.1          & 0.5 ± 0.0          & \textbf{545.1 ± 2.6}    & 53.7 ± 0.8          & 13.0 ± 0.9          & \multicolumn{1}{c|}{\textbf{72.0 ± 1.9}} & 54.5 ± 0.6          & 12.7 ± 0.5          & 71.2 ± 2.3          \\
\multicolumn{1}{c|}{}                         & \multicolumn{1}{c|}{\textbf{+REP}}   & 87.1 ± 0.1          & 99.4 ± 0.1          & 0.5 ± 0.0          & 543.3 ± 2.3             & \textbf{57.0 ± 0.3} & \textbf{17.4 ± 0.6} & \multicolumn{1}{c|}{71.9 ± 1.9}          & \textbf{56.2 ± 0.6} & \textbf{14.5 ± 0.2} & \textbf{71.8 ± 1.5} \\
\multicolumn{1}{c|}{}                         & \multicolumn{1}{c|}{\textbf{R-ROME}} & \textbf{92.1 ± 0.2} & 99.7 ± 0.0          & \textbf{0.5 ± 0.0} & \textbf{565.0 ± 0.9}    & 43.1 ± 0.9          & 4.6 ± 0.2           & \multicolumn{1}{c|}{68.7 ± 0.4}          & 43.1 ± 1.1          & 4.4 ± 0.3           & 68.7 ± 1.4          \\
\multicolumn{1}{c|}{}                         & \multicolumn{1}{c|}{\textbf{+REP}}   & 89.8 ± 0.3          & 99.7 ± 0.0          & 0.8 ± 0.2          & 562.0 ± 2.2             & \textbf{71.4 ± 0.6} & \textbf{51.2 ± 0.7} & \multicolumn{1}{c|}{\textbf{72.4 ± 0.2}} & \textbf{57.2 ± 0.9} & \textbf{25.2 ± 0.9} & \textbf{72.0 ± 1.7} \\
\multicolumn{1}{c|}{}                         & \multicolumn{1}{c|}{\textbf{EMMET}}  & \textbf{86.6 ± 1.4} & 99.7 ± 0.1          & \textbf{0.5 ± 0.0} & \textbf{563.8 ± 0.9}    & 33.0 ± 1.3          & 2.8 ± 0.3           & \multicolumn{1}{c|}{52.1 ± 2.6}          & 33.0 ± 1.1          & 2.8 ± 0.3           & 52.7 ± 3.9          \\
\multicolumn{1}{c|}{}                         & \multicolumn{1}{c|}{\textbf{+REP}}   & 84.7 ± 1.3          & 99.7 ± 0.1          & 0.7 ± 0.1          & 561.6 ± 2.0             & \textbf{66.7 ± 2.3} & \textbf{50.9 ± 2.2} & \multicolumn{1}{c|}{\textbf{59.4 ± 2.2}} & \textbf{50.0 ± 1.8} & \textbf{22.6 ± 1.8} & \textbf{59.7 ± 3.7} \\ \midrule
\multicolumn{1}{c|}{\multirow{8}{*}{\texttt{Mistral}}} & \multicolumn{1}{c|}{\textbf{ROME}}   & \textbf{97.2 ± 0.3} & 99.5 ± 0.1          & 1.6 ± 0.1          & 584.0 ± 2.5             & 49.2 ± 0.4          & 4.2 ± 0.4           & \multicolumn{1}{c|}{77.2 ± 0.1}          & 50.3 ± 0.9          & 4.1 ± 0.7           & 78.3 ± 0.9          \\
\multicolumn{1}{c|}{}                         & \multicolumn{1}{c|}{\textbf{+REP}}   & 93.1 ± 0.6          & 99.5 ± 0.1          & \textbf{1.5 ± 0.0} & \textbf{584.6 ± 1.0}    & \textbf{84.3 ± 0.8} & \textbf{74.1 ± 1.1} & \multicolumn{1}{c|}{\textbf{78.6 ± 0.6}} & \textbf{71.5 ± 1.5} & \textbf{40.8 ± 1.5} & \textbf{79.1 ± 1.4} \\
\multicolumn{1}{c|}{}                         & \multicolumn{1}{c|}{\textbf{MEMIT}}  & \textbf{94.1 ± 1.0} & 99.4 ± 0.1          & 1.4 ± 0.1          & \textbf{579.6 ± 3.1}    & 60.5 ± 0.8          & 12.4 ± 0.8          & \multicolumn{1}{c|}{\textbf{80.9 ± 2.2}} & 62.1 ± 1.2          & 11.9 ± 0.3          & \textbf{81.8 ± 1.2} \\
\multicolumn{1}{c|}{}                         & \multicolumn{1}{c|}{\textbf{+REP}}   & 90.2 ± 0.7          & 99.4 ± 0.1          & \textbf{1.3 ± 0.0} & 579.0 ± 1.6             & \textbf{68.9 ± 1.9} & \textbf{36.5 ± 2.4} & \multicolumn{1}{c|}{80.0 ± 1.9}          & \textbf{66.1 ± 1.7} & \textbf{25.9 ± 1.6} & 80.8 ± 1.6          \\
\multicolumn{1}{c|}{}                         & \multicolumn{1}{c|}{\textbf{R-ROME}} & \textbf{97.5 ± 0.2} & 99.6 ± 0.2          & 1.6 ± 0.1          & \textbf{585.5 ± 2.5}    & 50.1 ± 0.4          & 4.3 ± 0.3           & \multicolumn{1}{c|}{78.4 ± 0.8}          & 50.9 ± 0.7          & 4.4 ± 0.8           & 78.8 ± 1.1          \\
\multicolumn{1}{c|}{}                         & \multicolumn{1}{c|}{\textbf{+REP}}   & 93.3 ± 0.7          & 99.6 ± 0.2          & 1.6 ± 0.1          & 585.0 ± 4.3             & \textbf{84.9 ± 0.7} & \textbf{75.6 ± 0.7} & \multicolumn{1}{c|}{\textbf{79.3 ± 0.6}} & \textbf{72.0 ± 1.6} & \textbf{42.4 ± 1.5} & \textbf{80.3 ± 1.4} \\
\multicolumn{1}{c|}{}                         & \multicolumn{1}{c|}{\textbf{EMMET}}  & \textbf{95.9 ± 0.3} & 99.5 ± 0.1          & 1.7 ± 0.1          & 588.0 ± 0.9             & 41.8 ± 1.2          & 2.9 ± 0.5           & \multicolumn{1}{c|}{52.8 ± 4.2}          & 42.4 ± 1.4          & 2.7 ± 0.3           & 52.6 ± 5.4          \\
\multicolumn{1}{c|}{}                         & \multicolumn{1}{c|}{\textbf{+REP}}   & 90.2 ± 1.0          & 99.5 ± 0.1          & \textbf{1.6 ± 0.1} & \textbf{589.1 ± 1.8}    & \textbf{83.0 ± 0.6} & \textbf{76.1 ± 0.7} & \multicolumn{1}{c|}{\textbf{61.0 ± 3.2}} & \textbf{68.4 ± 1.5} & \textbf{44.8 ± 2.0} & \textbf{60.2 ± 4.2} \\ \midrule
\multicolumn{1}{c|}{\multirow{8}{*}{\texttt{Qwen2}}}   & \multicolumn{1}{c|}{\textbf{ROME}}   & \textbf{98.3 ± 0.1} & \textbf{98.9 ± 0.2} & 2.0 ± 0.0          & 562.1 ± 3.7             & 53.8 ± 0.2          & 11.0 ± 0.5          & \multicolumn{1}{c|}{76.8 ± 0.4}          & 55.9 ± 0.1          & 11.0 ± 0.9          & 77.5 ± 1.1          \\
\multicolumn{1}{c|}{}                         & \multicolumn{1}{c|}{\textbf{+REP}}   & 97.0 ± 0.4          & 97.4 ± 0.0          & 2.0 ± 0.0          & \textbf{568.8 ± 2.9}    & \textbf{64.2 ± 0.4} & \textbf{42.2 ± 1.0} & \multicolumn{1}{c|}{\textbf{80.6 ± 0.5}} & \textbf{61.7 ± 0.8} & \textbf{36.0 ± 0.4} & \textbf{79.2 ± 1.0} \\
\multicolumn{1}{c|}{}                         & \multicolumn{1}{c|}{\textbf{MEMIT}}  & \textbf{95.4 ± 0.2} & \textbf{98.2 ± 0.1} & \textbf{1.5 ± 0.0} & 573.9 ± 5.8             & 62.4 ± 0.3          & 22.2 ± 0.3          & \multicolumn{1}{c|}{88.8 ± 1.6}          & 65.2 ± 0.3          & 22.1 ± 0.8          & \textbf{89.2 ± 0.5} \\
\multicolumn{1}{c|}{}                         & \multicolumn{1}{c|}{\textbf{+REP}}   & 94.2 ± 0.3          & 97.2 ± 0.1          & 1.6 ± 0.1          & \textbf{576.4 ± 4.1}    & \textbf{78.7 ± 0.9} & \textbf{51.9 ± 2.0} & \multicolumn{1}{c|}{88.8 ± 1.5}          & \textbf{74.8 ± 0.7} & \textbf{41.2 ± 2.4} & 89.0 ± 0.4          \\
\multicolumn{1}{c|}{}                         & \multicolumn{1}{c|}{\textbf{R-ROME}} & \textbf{98.2 ± 0.2} & \textbf{98.8 ± 0.1} & \textbf{2.0 ± 0.0} & 571.5 ± 2.2             & 54.2 ± 0.3          & 11.5 ± 0.5          & \multicolumn{1}{c|}{77.2 ± 1.3}          & 56.7 ± 0.1          & 12.0 ± 0.8          & 76.7 ± 1.5          \\
\multicolumn{1}{c|}{}                         & \multicolumn{1}{c|}{\textbf{+REP}}   & 96.0 ± 0.2          & 97.4 ± 0.1          & 2.2 ± 0.3          & \textbf{575.5 ± 3.3}    & \textbf{66.3 ± 1.0} & \textbf{45.3 ± 0.4} & \multicolumn{1}{c|}{\textbf{80.0 ± 2.1}} & \textbf{63.3 ± 0.7} & \textbf{38.8 ± 0.5} & \textbf{78.9 ± 2.5} \\
\multicolumn{1}{c|}{}                         & \multicolumn{1}{c|}{\textbf{EMMET}}  & \textbf{94.6 ± 0.2} & \textbf{97.7 ± 0.3} & \textbf{1.5 ± 0.0} & 570.6 ± 4.2             & 58.5 ± 0.5          & 17.7 ± 0.4          & \multicolumn{1}{c|}{\textbf{75.2 ± 1.8}} & 60.3 ± 0.7          & 18.9 ± 0.5          & 75.6 ± 0.9          \\
\multicolumn{1}{c|}{}                         & \multicolumn{1}{c|}{\textbf{+REP}}   & 91.4 ± 0.3          & 91.7 ± 0.1          & 1.8 ± 0.2          & \textbf{574.5 ± 4.6}    & \textbf{78.2 ± 0.5} & \textbf{70.7 ± 0.5} & \multicolumn{1}{c|}{77.2 ± 1.8}          & \textbf{75.0 ± 0.9} & \textbf{67.6 ± 1.0} & \textbf{77.2 ± 0.8} \\ \bottomrule
\end{tabular}%
}
\end{table*}

\begin{figure*}[t]
    \centering
    \includegraphics[width=1\textwidth]{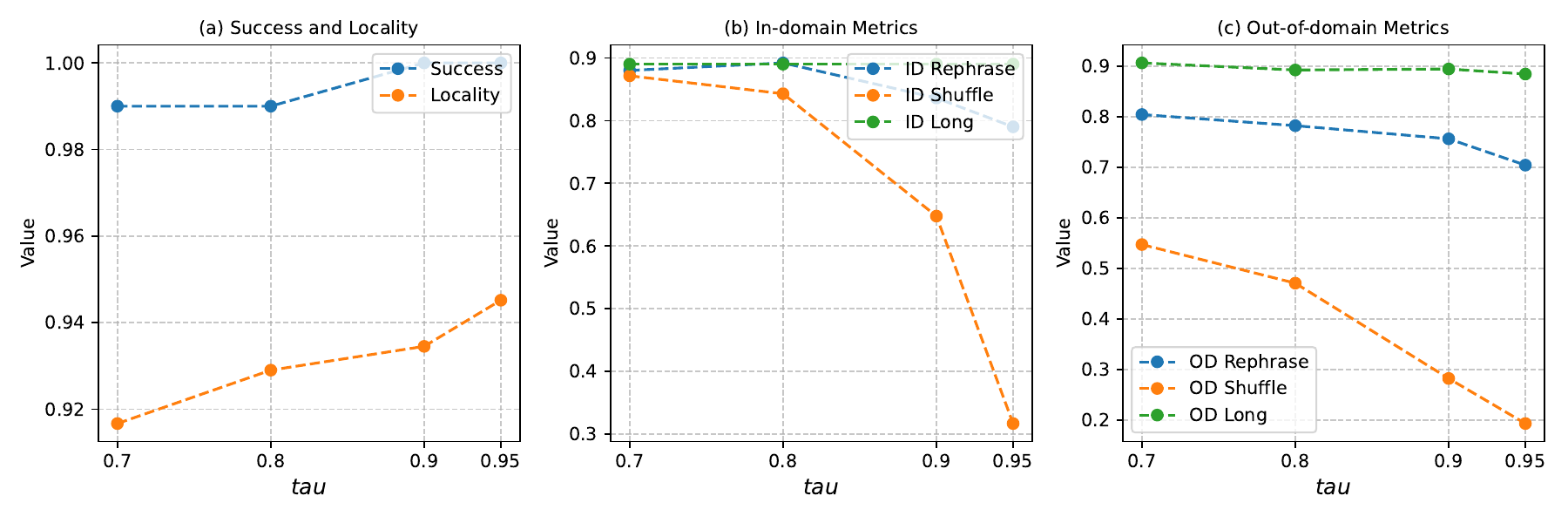}
    \caption{Hyper-parameter study of $\tau$ on validation set.}
    \label{ablation:tau}
  \end{figure*}
  
\section{Ablation Study}
\begin{figure*}[t]
    \centering
    \includegraphics[width=1\textwidth]{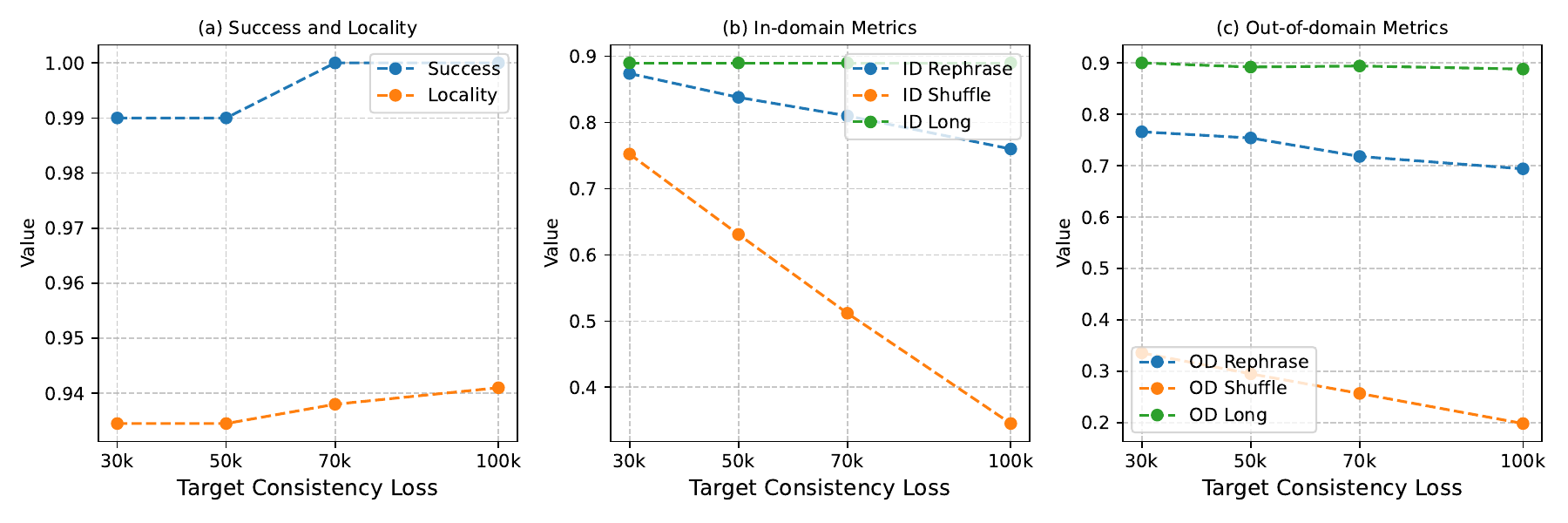}
    \caption{Hyper-parameter study of consistency loss weight $\alpha$ on validation set.}
    \label{ablation:target_loss}
  \end{figure*}

The gate threshold \( \tau \) and consistency loss weight \( \alpha \) significantly influence REP's performance, as discussed in Section \ref{sec:ablation}. Empirical analysis (Figures \ref{ablation:tau} and \ref{ablation:target_loss}) demonstrates that increasing \( \tau \) and \( \alpha \) improves locality preservation and edit success rates. However, robustness metrics initially plateau before deteriorating with further parameter escalation, underscoring the need to balance precision against generalization. This trade-off analysis justifies our selection of \( \tau = 0.9 \) and \( \alpha = 1\text{e+5} \), which optimally reconcile competing objectives across experiments.

\section{Case Visualization}
      \begin{figure}[t]
      \centering
      \includegraphics[width=\linewidth]{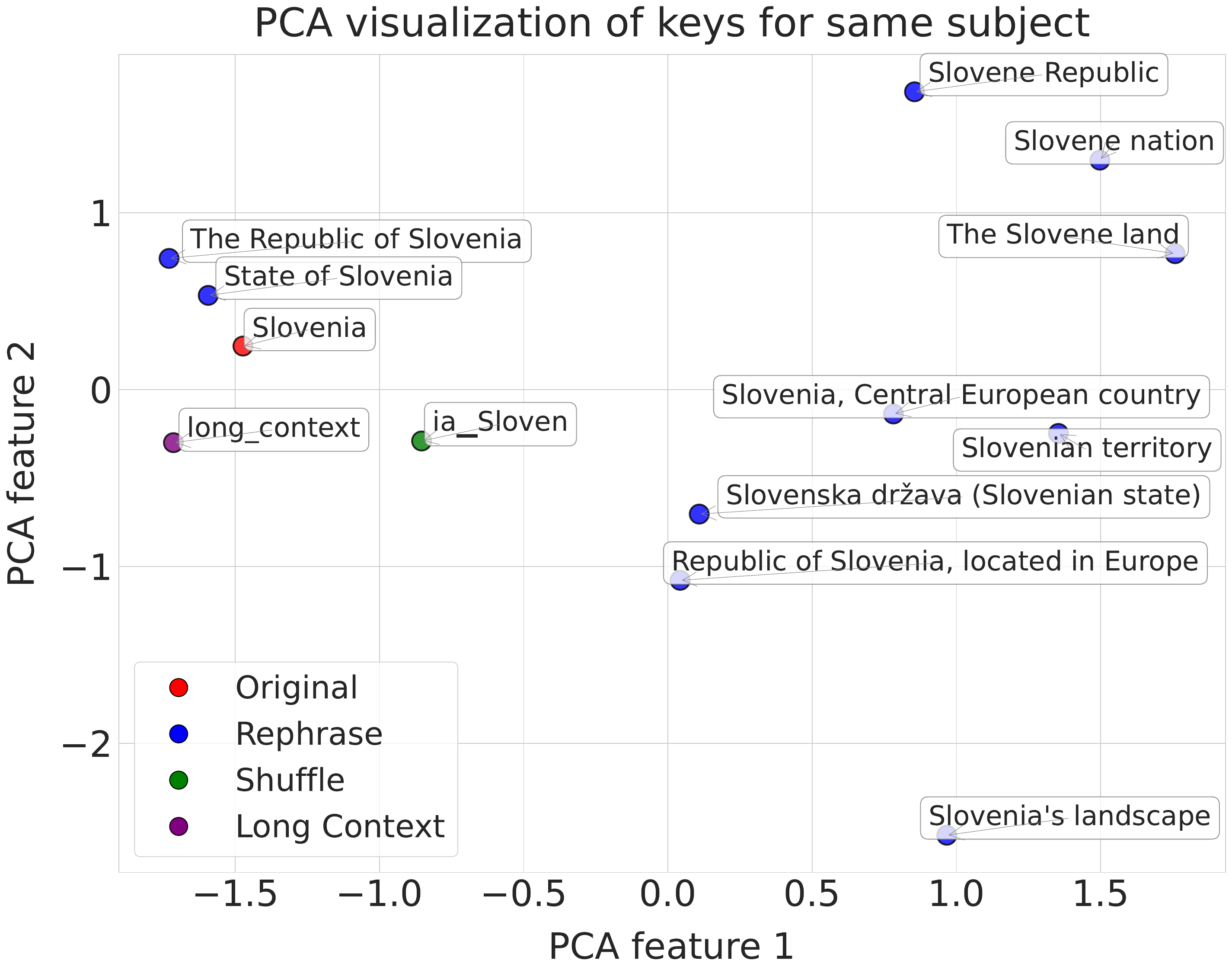}
      \caption{A visualization for key representations of rephrases for `Slovenia' in LLaMA2-7B.}
      \label{fig:case}
    \end{figure}

Figure \ref{fig:case} (right) provides a visualization of representations for the subject {'Slovenia'} after three types of perturbations, reduced to two dimensions using Principal Component Analysis (PCA). This visualization corroborates our previous findings:
(1) \emph{Context sensitivity:} Long irrelevant context induces a slight shift in the representation, indicating contextual influence on subject encoding.
(2) \emph{Rephrase variability:} Rephrased versions of the subject sometimes cluster close to the original representation, while at other times they are distant. 
(3) \emph{Order dependence:} Shuffling the word order results in substantial deviations from the original representation. This observation highlights the model's sensitivity to word order, even when the constituent tokens remain unchanged.

When the edited key has near-zero or negative similarity with other keys, based on Lemma \ref{lemma:generalize} it becomes virtually impossible for the edited value to be retrieved, potentially compromising the robustness of the edit.

\section{Analyzing Value Distributions}
\paragraph{Loud Voices.}

\begin{figure}[h]
    \centering
    \captionsetup{skip=2pt}
    \includegraphics[width=0.8\linewidth]{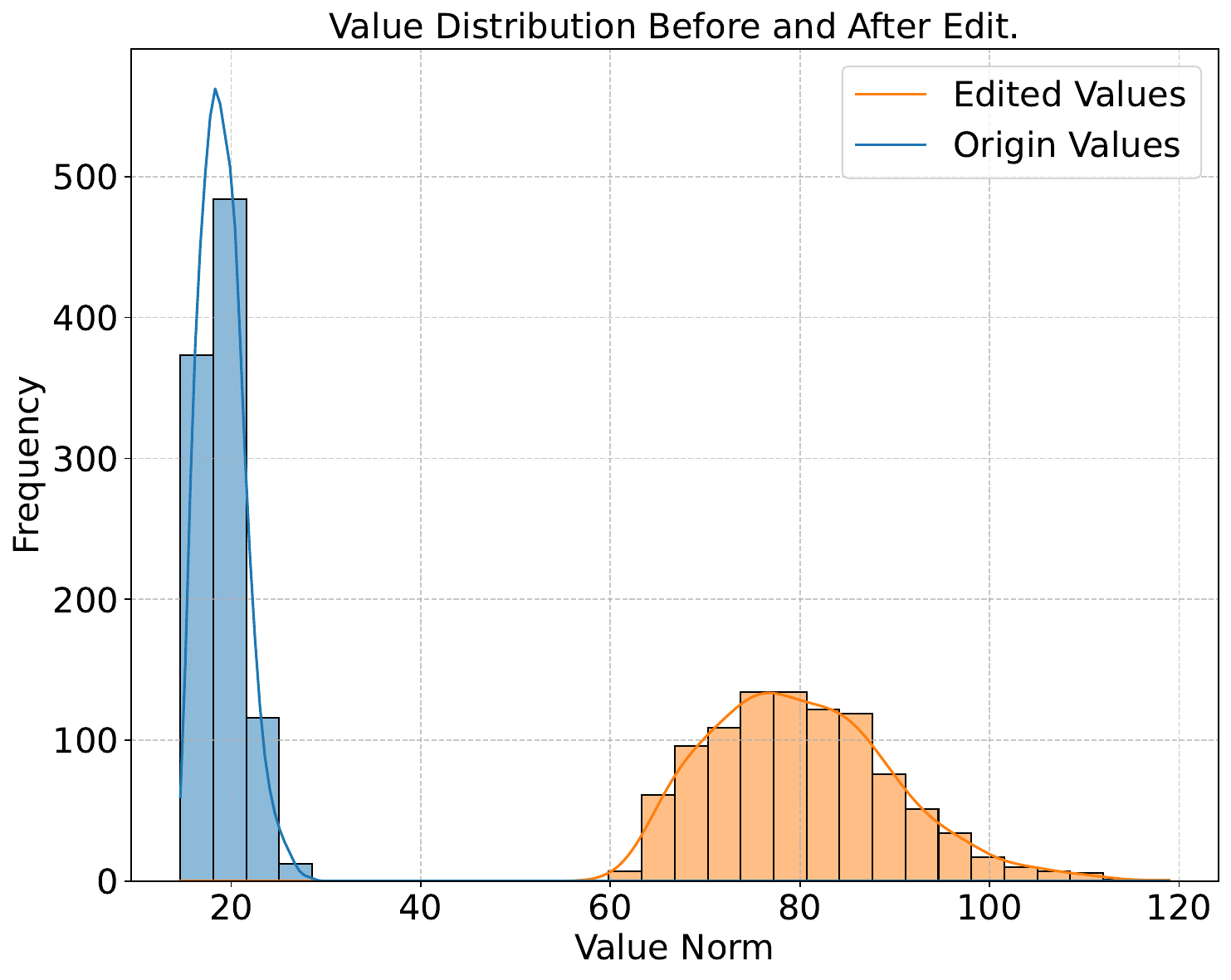}
    \caption{Values before and after edit with ROME.}
    \label{fig:values}
    \vspace{-20pt}
\end{figure}

In Figure \ref{fig:values}, we present the distribution of values before and after edits, using LLaMA-2 7B and ROME. The results demonstrate that post-edit values exhibit significantly larger L2 norms compared to pre-edit values. This observation aligns with our findings in Lemma \ref{lemma:value_bound} and \ref{lemma:generalize}, which suggest that edited values must be sufficiently large to effect changes on the current key and influence distant keys.

However, this increase in value magnitude, while necessary for effective editing, presents potential challenges. As indicated by Lemma \ref{lemma:spec} and our previous analysis, these 'loud' values may inadvertently affect unrelated keys, particularly those that are proximal in the representation space to the one being edited. This observation highlights a tension between achieving targeted edits and avoiding unintended consequences in the model's broader knowledge representation.

\paragraph{Summary.}
Our findings collectively suggest that the inner representations of large language models (LLMs) may not serve as reliable keys for editing purposes. The observed variability in key similarities, even among semantically equivalent subjects, coupled with the necessity for large-magnitude value changes, poses significant challenges for precise and controlled model editing. These issues can lead to unintended effects on unrelated parts of the model's knowledge and compromise the specificity of edits. Furthermore, the sensitivity of representations to word order and context underscores the instability of using these internal states as edit targets. 
These limitations motivate us to explore alternative approaches, particularly the concept of \emph{branching a separate path for keys}. 
By creating a dedicated pathway for key representations, we may achieve more stable and controllable edit targets, potentially mitigating the issues of representation variability and unintended side effects observed when directly manipulating the model's inner representations.



\end{document}